\documentclass[dvipsnames]{article}

\usepackage[final]{openbmb_2025}

\usepackage{amsmath}
\usepackage{amssymb}
\usepackage{mathtools}
\usepackage{amsthm}
\usepackage{multirow}
\usepackage{makecell}
\usepackage{subcaption}
\usepackage{wrapfig}
\usepackage{graphicx}

\usepackage{pifont}
\usepackage{fvextra}
\usepackage{xspace}

\usepackage[utf8]{inputenc} 
\usepackage[T1]{fontenc}    
\usepackage{url}            
\usepackage{booktabs}       
\usepackage{amsfonts}       
\usepackage{nicefrac}       
\usepackage{microtype}      

\usepackage{color, soul}
\usepackage[most,breakable]{tcolorbox}
\usepackage{textgreek}

\usepackage[labelfont=bf]{caption}
\usepackage{enumitem}
\usepackage{sectsty}
\usepackage{pgfplots}
\usepackage{fancyhdr}
\usepackage{xcolor} 
\renewcommand{\headrulewidth}{1pt}
\makeatletter
\def\headrule{{\if@fancyplain\let\headrulewidth\plainheadrulewidth\fi
\hrule\@height\headrulewidth\@width\textwidth \vskip-\headrulewidth}}
\makeatother

\definecolor{BMBDarkBlue}{HTML}{315EFE}
\definecolor{BMBLightBlue}{HTML}{00D3ED}
\usepackage[colorlinks, linkcolor=BMBDarkBlue, anchorcolor=BMBDarkBlue, citecolor=BMBDarkBlue, urlcolor=BMBDarkBlue]{hyperref}
\usepackage{cleveref}
\usepackage{threeparttable}

\sectionfont{\color{BMBDarkBlue}\fontfamily{zi4}\selectfont}
\subsectionfont{\color{BMBDarkBlue}\fontfamily{zi4}\selectfont}
\subsubsectionfont{\color{BMBDarkBlue}\fontfamily{zi4}\selectfont}

\newtcolorbox{mytheorem}{
  colback=gray!5,       
  colframe=gray!80,     
  boxrule=0.5pt,        
  arc=4pt,              
  left=4pt,             
  right=4pt,            
  top=4pt,              
  bottom=4pt,           
}

\newcommand{\fancyheadname}{\textit{\textbf{\modelname{}}}}

\title{\modelname{}: Ultra-Efficient LLMs on End Devices}

\author{%
\\
\textbf{\large{MiniCPM Team}}
\vspace{0em}
}

\def\huggingface{\raisebox{-1.5pt}{\includegraphics[height=1.05em]{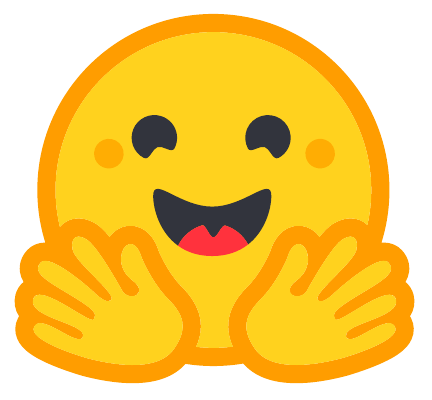}}}
\def\github{\raisebox{-1.5pt}{\includegraphics[height=1.05em]{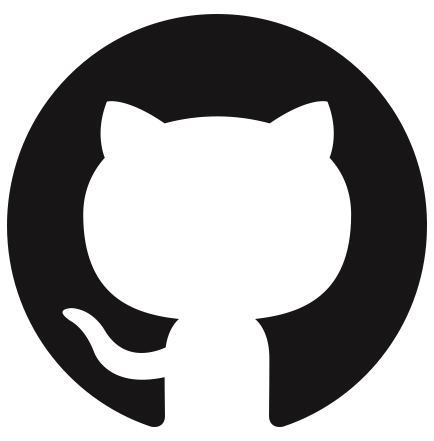}}}
\newcommand{\modelname}[0]{MiniCPM4}

\newcommand{\surveyagent}{\modelname{}-Survey\xspace}

\usepackage{geometry}
\usepackage{algorithm}
\usepackage{algpseudocode}
\usepackage{setspace}

\makeatletter
\renewcommand{\ALG@beginalgorithmic}{\small}
\makeatother



\begin{document}

\maketitle
\thispagestyle{fancy} 

\begin{abstract}
This paper introduces \modelname{}, a highly efficient large language model (LLM) designed explicitly for end-side devices.
We achieve this efficiency through systematic innovation in four key dimensions: model architecture, training data, training algorithms, and inference systems. 
Specifically, in terms of model architecture, we propose InfLLM v2, a trainable sparse attention mechanism that accelerates both prefilling and decoding phases for long-context processing.
Regarding training data, we propose UltraClean, an efficient and accurate pre-training data filtering and generation strategy, 
and UltraChat v2, a comprehensive supervised fine-tuning dataset. These datasets enable satisfactory model performance to be achieved using just $8$ trillion training tokens. 
Regarding training algorithms, we propose ModelTunnel v2 for efficient pre-training strategy search, and improve existing post-training methods by introducing chunk-wise rollout for load-balanced reinforcement learning and data-efficient tenary LLM, BitCPM.
Regarding inference systems, we propose CPM.cu that integrates sparse attention, model quantization, and speculative sampling to achieve efficient prefilling and decoding.
To meet diverse on-device requirements, \modelname{} is available in two versions, with 0.5B and 8B parameters, respectively. Furthermore, we construct a hybrid reasoning model, \modelname{}.1, which can be used in both deep reasoning mode and non-reasoning mode.
Evaluation results demonstrate that \modelname{} and \modelname{}.1 outperform similar-sized open-source models across benchmarks, with the 8B variants showing significant speed improvements on long sequence understanding and generation.
\end{abstract}

\begin{figure}[h]
    \centering
    \includegraphics[width=0.9\linewidth]{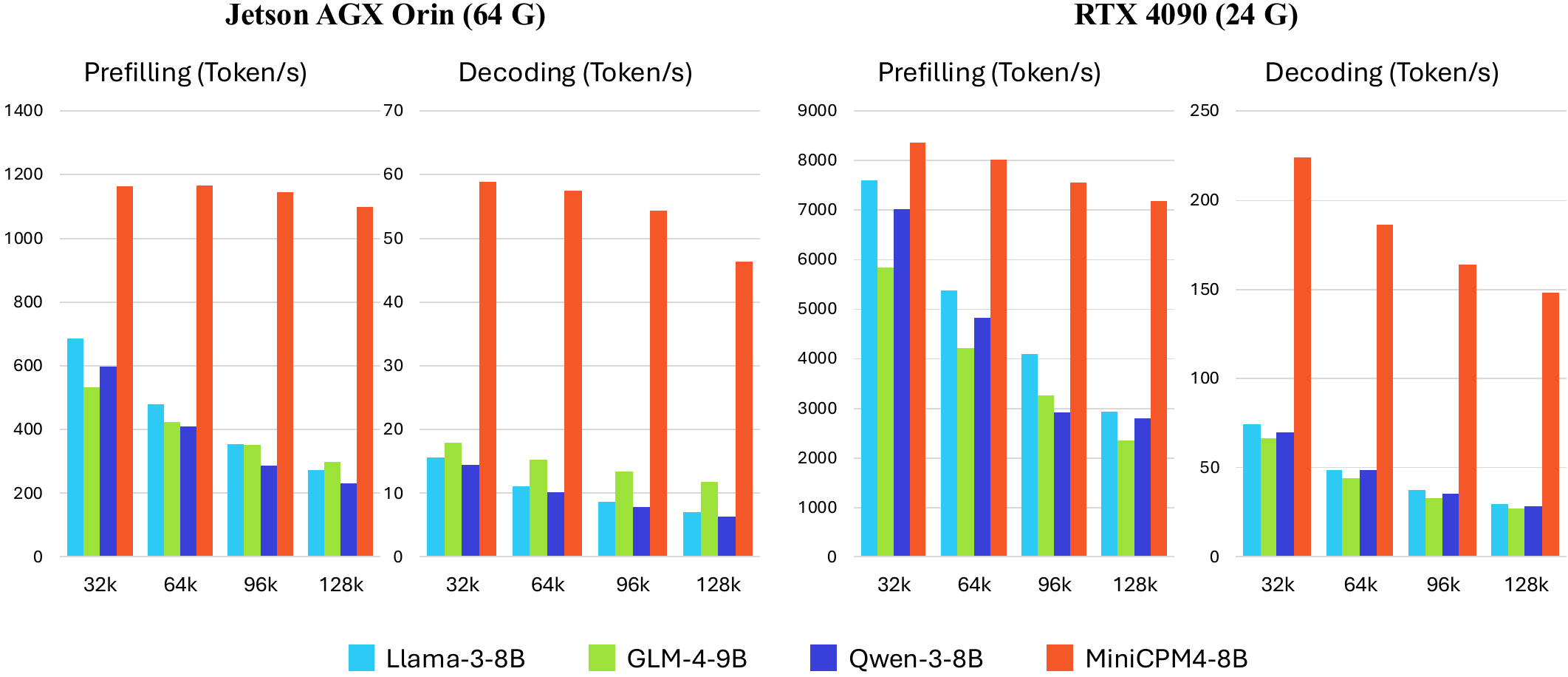}
    \caption{Inference Speed Evaluation on end-side GPUs.}
    \label{fig:efficiency-eval}
\end{figure}

\newpage
{
  \hypersetup{linkcolor=RoyalBlue, linktoc=page}
  \tableofcontents
}

\newpage

\section{Introduction}  





Large language models (LLMs)~\citep{gpt3,gpt4}, also known as foundation models~\citep{foundation-model}, have become the core driving force in the field of artificial intelligence (AI)~\citep{PTMs-survey-qiu,PTMs-survey}. These large models exhibit impressive abilities to handle diverse tasks, from helpful chatbot systems~\citep{instruct-gpt} to complex reasoning systems~\citep{o1,DeepSeek-r1}, significantly enhancing the quality and efficiency of human-machine interaction. However, as the size of models continues to expand~\citep{scaling-law,chinchilla}, the requirement for computational resources grows exponentially, resulting in these models being primarily deployed on cloud servers and accessed through API interfaces.

The development of LLMs is currently facing an important trend toward miniaturization and increased efficiency. From the LLM application perspective, efficient models can reduce deployment costs and expand application scenarios, particularly in environments with limited computational resources such as end-side devices and mobile terminals~\citep{apple-intelligence,GPT-4o-mini}. From the technical development perspective, as model sizes continue to grow, improving computational efficiency becomes crucial for overcoming performance bottlenecks with limited resources~\citep{deepseek-v3}. Therefore, efficient model architectures and algorithms that maintain model capabilities while minimizing computational requirements are of considerable theoretical and practical significance.

Aligning with this move towards more efficient LLMs, our team has consistently concentrated on building efficient end-side MiniCPM models~\citep{minicpm,minicpm-v}. In this paper, we further boost model efficiency through systematic innovation in four key dimensions: model architecture, training data, training algorithms, and inference systems. Based on these advancements, we successfully develop \modelname{}, an $8$B LLM capable of efficient computation on edge-side chips. Notably, compared to the effective LLM Qwen3-8B, \modelname{} achieves comparable performance using $22\%$ of its training data, while simultaneously demonstrating a 7-fold speed improvement in processing 128K-length documents on end-side devices.  

Specifically, \modelname{} is featured with the following technologies to improve computational efficiency while maintaining model capabilities:

\textbf{Model Architecture: Trainable Sparse Attention}\quad
With the widespread application of LLMs in long-context processing~\citep{deepresearch,swe-bench} and the drive for deep reasoning capabilities~\citep{DeepSeek-r1,o1}, the need for LLMs to comprehend and generate long sequences has become increasingly critical. However, the computational and memory demands of self-attention mechanisms pose significant challenges for efficiently processing lengthy documents on end-side devices. We propose a sparse attention architecture, enabling efficient long-context processing while maintaining model performance.
\begin{itemize}[leftmargin=*]
    \item \textit{InfLLM v2 -- Trainable Sparse Attention Capable of Prefilling and Decoding Acceleration}:
    Building upon our dynamic sparse attention architecture, InfLLM~\citep{xiao2024infllm}, we introduce InfLLM v2, which features efficient kernel design and end-to-end specialized training. Its kernel facilitates token-level sparse attention computation at the query level, yielding significant speed improvements in both long-context prefilling and decoding phases. Additionally, we develop a specialized training framework that further enhances the sparsity of the attention mechanism and improves long-context processing capabilities.
\end{itemize}

\textbf{Training Data: Efficient Training Enhanced by High-Quality Data}\quad
High-quality data is crucial for enhancing the capability density of LLMs~\citep{xiao2024densing}. While massive internet corpora offer abundant training signals, they inevitably contain noise that leads to suboptimal performance. Building upon existing approaches to data cleaning and filtering~\citep{penedo2024fineweb}, we introduce an efficient iterative data cleaning strategy, UltraClean, which yields UltraFineWeb~\citep{ultrafineweb}, a high-quality knowledge-intensive dataset. Additionally, recognizing the scarcity of reasoning-intensive data, we conduct large-scale data synthesis specifically for mathematics and coding. These approaches enable us to develop a model with satisfactory performance using only $8$ trillion tokens of pre-training data.
\begin{itemize}[leftmargin=*]
    \item \textit{UltraClean -- High-Quality Pre-Training Data Filtering:} 
    We develop an efficient and effective data filtering strategy, UltraClean, which features an efficient verification strategy and an efficient quality classifier.
    Specifically, in contrast to conventional approaches that verify data quality by training LLMs from scratch using candidate corpora, our proposed efficient verification strategy leverages a nearly-trained LLM as a foundation. We incorporate candidate corpora during the final training steps and utilize the resulting performance improvement as a metric for assessing data quality. This verification strategy significantly enhances evaluation efficiency while maintaining quality assessment accuracy.
    Based on our efficient verification strategy, we can impartially select high-quality seed data for classifier training. 
    Building upon the assumption that ``high-quality seed data is beneficial for LLM training'', we develop and optimize the strategy for selecting classifier training seeds and recipes, while carefully curating balanced sets of both positive and negative samples to ensure classifier quality and robustness.
    We apply the proposed data filtering pipeline to the FineWeb~\citep{penedo2024fineweb} and Chinese FineWeb~\citep{chinese_fineweb} datasets, collectively termed \textit{\textbf{UltraFineWeb}}.
    This pipeline not only improves filtering efficiency, classifier quality, and robustness, but also significantly reduces experimental and inference costs.
    \item \textit{UltraChat v2 -- High-Quality Supervised Fine-Tuning Data Generation:}
    Building upon UltraChat~\citep{ding2023enhancing}, we introduce a high-quality dialogue construction strategy that enables the efficient generation of reasoning-intensive data for LLM training and evaluation.
    Specifically, in contrast to conventional instruction-tuning datasets that prioritize coverage or surface-level diversity, our proposed strategy focuses on multi-turn interactions with deep reasoning, contextual consistency, and task complexity. 
    We leverage a combination of expert models and prompt engineering to produce dialogues that challenge LLMs across a range of reasoning types, including multi-hop inference, commonsense reasoning, and domain-specific problem-solving. This approach not only enhances data quality but also supports the creation of robust and challenging benchmarks for fine-tuning and evaluation.
    Based on this reasoning-intensive generation pipeline, we construct high-quality seed data and optimize the design of prompts, dialogue structures, and quality control mechanisms. We further adopt a dual-stage filtering process that combines automated verification with selective human review to ensure response accuracy, coherence, and diversity.
    We apply the proposed pipeline to generate UltraChat v2, a reasoning-intensive extension of existing instruction-tuning corpora.
    This pipeline not only improves data generation quality and filtering efficiency, but also significantly enhances the reasoning capabilities of LLMs fine-tuned on this data.
\end{itemize}

\textbf{Training Algorithms: Multi-Dimensional Training Optimization Strategies}\quad
The scaling law of LLMs indicates that performance improves with increased training volume~\citep{scaling-law}. Reducing training costs is therefore essential for sustainable model scaling. In \cite{minicpm}, we develop ModelTunnel to search for the optimal training strategy by conducting a series of experiments on small-scale models. In developing \modelname{}, we further improve the search accuracy and introduce ModelTunnel v2.
For efficient post-training mechanism, we employ a load-balanced rollout strategy for the reinforcement learning (RL) process, which can make better use of the computational resources during the rollout process. Besides, to reduce the storage requirements, we introduce a ternary LLM, BitCPM4.
\begin{itemize}[leftmargin=*]
    \item \textit{ModelTunnel v2 -- Efficient Training Strategy Search:}
    Following our ModelTunnel~\citep{minicpm}, we develop ModelTunnel v2, which advances in two aspects: 
    (1) Improved performance indicator: Due to the emergenet abilities~\citep{emergent-ability}, previous predictable scaling methods cannot effectively predict the performance of downstream tasks. In \modelname{}, we construct ScalingBench~\citep{xiao2024densing} and establish the relationship between the loss of ScalingBench and downstream performance. Therefore, instead of using language model loss as the performance indicator for predictable scaling, we can use ScalingBench as the performance indicator, which can improve the hyper-parameter searching effectiveness.
    (2) Search effectiveness validation: Utilizing Scaling-Bench~\citep{xiao2024densing}, we systematically validate the effectiveness of the identified parameters. Our experimental results demonstrate that the maximal-update-parameterization ($\mu$P)~\citep{yang2022tensor}  combined with our hyperparameter search achieves performance comparable to state-of-the-art industry methods and our method can significantly reduce the searching costs. 
    \item \textit{Chunk-wise Rollout -- Load-Balanced Reinforcement Learning:} We implement a chunk-wise rollout strategy to improve RL training efficiency, which limits the maximum output token budget for each rollout phase and resumes the generation of incomplete trajectories in subsequent iterations, significantly reducing idle computations caused by lengthy trajectories. Additionally, to address the instability introduced by the chunk-wise rollout strategy, we incorporate several stabilization techniques, including KL loss, dual-clip, chunk-level importance sampling, and garble filter. Collectively, these enhancements ensure stable and efficient scaling of long chain-of-thought (CoT) RL training.
    \item \textit{BitCPM4 -- Quantization-Aware Training for Ternary LLMs:} We design a two-stage training framework that substantially reduces quantization-aware training (QAT) costs by initializing the quantization phase with our pre-trained high-precision model. Integrated with ModelTunnel v2, our method achieves comparable performance to existing QAT approaches while using 10$\times$ fewer training tokens. For those extremely resource-limited devices, we adapt \modelname{} to the ternary version BitCPM4 and show promising results.
    \item \textit{Efficient Training Engineering:} Inspired by \cite{deepseek-v3}, we implement both the multi-token prediction training objective~\citep{gloeckle2024better} and the FP-8 mixed-precision training framework. Multi-token prediction can introduce more intensive supervision signals and make the additional head achieve a higher acceptance length in speculative sampling. FP-8 mixed-precision training can make full use of the computational power of our GPU clusters.
\end{itemize}

\textbf{Inference Systems: High-Performance Inference Framework for End-Side Devices}\quad
End-side devices have limited computational power and storage resources. To fully utilize these devices, we customize a high-performance inference framework. 
\begin{itemize}[leftmargin=*]
    \item \textit{CPM.cu -- Lightweight and Efficient CUDA Inference Framework:} We first develop a lightweight inference framework featuring static memory management, kernel fusion, and efficient speculative sampling implementation, achieving efficient prefilling and decoding speed. Building upon this framework, we integrate efficient sparse attention kernels for InfLLM v2, further improve speculative drafting speed with FR-Spec~\citep{zhao2025fr}, introduce the more effective prefix-aware quantization method P-GPTQ, and investigate the combined effect of speculative sampling and quantization through SpecMQuant~\citep{zhang2025specmqaunt}.
    \item \textit{ArkInfer -- Cross-Platform Deployment Framework:} To address the challenge of deploying LLMs across diverse hardware platforms, we design ArkInfer with a unified executor-based architecture and adaptive backend interfaces. We integrate multiple inference frameworks (NeuroPilot, Genie, RK-LLM, TensorRT-LLM, and llama.cpp) through standardized APIs and employ advanced optimization techniques like typical speculative sampling and quantization methods. This design enables seamless cross-platform deployment with native multimodal capabilities, flexible sampling strategies, and comprehensive performance evaluation tools, significantly simplifying the integration of MiniCPM models across various end-side devices beyond NVIDIA chips.
\end{itemize}

Based on the above techniques, we build \modelname{} with two parameter versions: 0.5B and 8B, each with general and deep inference variants. During pre-training, we train the 8B model on 8.3T high-quality tokens. We adopt the warmup-stable-decay (WSD) learning rate scheduler~\citep{minicpm}, allocating 7T tokens for the warmup and stable phases and 1.3T tokens for the annealing phase. Following this, we conduct long-context pre-training to extend the context window of \modelname{} from 4K to 128K tokens. Subsequently, we perform supervised fine-tuning (SFT) post-training to enable the model to follow user instructions. To further develop a hybrid reasoning model, \modelname{}.1, we utilize long CoT data for SFT and implement reinforcement learning with mathematics and coding tasks. \modelname{}.1 is trained based on \modelname{} with improved post-training corpus. \modelname{}.1 is a hybrid model, and can be used in both reasoning and non-reasoning modes.
We evaluate \modelname{} and \modelname{}.1 on a series of widely-used benchmarks. \modelname{} and \modelname{}.1 outperform typical baselines with similar parameter sizes and becomes one of the most effective and efficient open-source LLMs.

Finally, based on \modelname{}, we develop three applications to present the effectiveness of \modelname{} and explore advanced technology, including trustworthy survey generation and tool use with model context protocol (MCP). All three applications require the model to be able to generate long sequences with high coherence and logicality, invoke complex functions to obtain external resources, and write creatively. The results show that \modelname{} presents promising results on these applications, and we encourage the community to explore more interesting applications based on our \modelname{}.

\section{Efficient Architecture and Pre-training} 
In this section, we introduce the model architecture and training algorithms applied in \modelname{}, which features efficient sparse attention layers and an efficient training pipeline. In this section, we will first introduce the details of our proposed InfLLM v2, supporting efficient long-context processing. Specifically, InfLLM v2 enables \modelname{} to achieve comparable long-context processing ability with the full attention mechanism with $81\%$ attention sparsity. Then, we describe the pre-training data management methods, enabling efficient knowledge learning. Finally, we present our pre-training pipeline, including ModelTunnel v2 for hyperparameter search as well as engineering for pre-training objective and long-context extension. All these methods enable \modelname{}-8B to achieve comparable results with Qwen3-8B using only $22\%$ of the pre-training tokens of Qwen3-8B.

\subsection{InfLLM v2: Trainable Sparse Attention for Prefilling and Decoding}
Following most open-source LLMs, we adopt Transformer~\citep{transformer} as our basic architecture. In consideration of the emerging needs to process long sequences, many efforts have been devoted to designing a training-free sparse attention mechanism to dynamically select relevant context tokens for long-context processing~\citep{stream-llm,jiangminference,xu2025xattention,zhang2025spargeattn}. These models can only be applied in prefilling acceleration due to their unsatisfactory sparsity. 

Recently, MoBA~\citep{lu2025moba} and NSA~\citep{yuan2025native} apply sparse attention in the pre-training stage to improve model performance. However, MoBA utilizes the design of query blocks, which prevents it from achieving acceleration during the decoding phase. Besides, according to our observations, the relevant contexts between adjacent tokens usually vary greatly. Therefore, forcing adjacent tokens to share the same context may lead to sub-optimal performance, and at the same time, the sparsity of attention cannot be improved. 
NSA introduces three different attention components to capture long-distance information. The three attention components introduce additional parameters, which will lead to increased computational overhead for short sequences and threefold key-value storage costs for pre-training.

In this section, based on our previous sparse attention model, InfLLM~\citep{xiao2024infllm}, we design a trainable sparse attention, InfLLM v2, to reduce the computation and memory access costs for both prefilling and decoding phrases. InfLLM v2 does not introduce additional parameters for attention output and will not influence the inference for short sequences. Besides, we propose an efficient Top-K context block selection method, which can reduce $60\%$ computational costs compared with NSA.
In the following paragraphs, we will introduce the algorithm design of InfLLM v2.

\begin{figure*}
    \centering
    \includegraphics[width=0.9\linewidth]{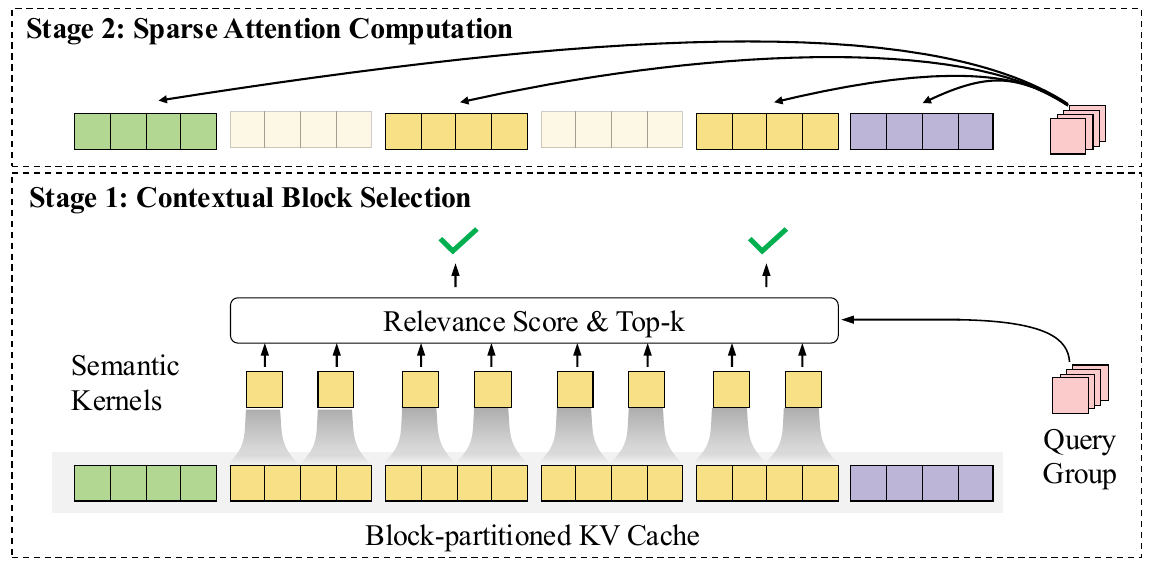}
    \caption{The illustration of InfLLM v2. Each query group selects parts of key-value blocks for attention computation, where the initial tokens and local tokens in the sliding window are always selected.}
    \label{fig:infllm-v2}
    \vspace{-1em}
\end{figure*}

\subsubsection{Overall Framework of InfLLM v2}
In a sparse attention mechanism, we only select parts of context tokens for attention computation. Following InfLLM, we split the key-value cache into block-level units, and each query token will select blocks with the highest relevance scores for attention. 
Formally, in each attention layer, the input is a sequence of hidden vectors, with each vector representing the contextual information of a token. Given an input sequence $\mathbf{X} = \{\mathbf{x}_1, \mathbf{x}_2, \ldots, \mathbf{x}_l\}$, the self-attention mechanism first maps this sequence to query, key, and value vectors $\mathbf{Q} =\{\mathbf{q}_1, \mathbf{q}_2, \ldots, \mathbf{q}_l\}, \mathbf{K} =\{\mathbf{k}_1, \mathbf{k}_2, \ldots, \mathbf{k}_l\}, \mathbf{V} =\{\mathbf{v}_1, \mathbf{v}_2, \ldots, \mathbf{v}_l\}$. Here, $l$ is the length of the input sequence. In a dense attention mechanism, each token needs to attend all preceding tokens, which can be expressed as $\mathbf{o}_i = \text{Attention}\left(\mathbf{q}_i, \left(\mathbf{K}_{1:i}, \mathbf{V}_{1:i}\right)\right)$. In contrast, a sparse attention mechanism selectively attends to only a subset of relevant tokens.
We partition the key-value cache into blocks to avoid fine-grained token-level relevance computation and memory access, thereby improving the efficiency of sparse attention. Specifically, InfLLM v2 divides the key-value cache into equal-sized blocks, with each block containing $m$ tokens.
The key-value cache is thus partitioned into blocks $\mathcal{B} = \{\mathbf{B}_0, \ldots, \mathbf{B}_{\lfloor\frac{l}{m}\rfloor - 1}\}$, where $\mathbf{B}_j = \left(\mathbf{K}_{jm:(j+1)m}, \mathbf{V}_{jm:(j+1)m}\right)$. Here, $m$ is the block size for key-value blocks.

The sparse attention computation of InfLLM v2 consists of two stages. In the first stage, we dynamically select relevant blocks from $\mathcal{B}$ based on the query token $\mathbf{q}_i$. To this end, we need to compute the relevance score $r_{\text{block}}(\mathbf{q}_i, \mathbf{B}_j)$ between the query token $\mathbf{q}_i$ and each block, and then select the blocks with the highest relevance scores. In the second phase, based on the blocks selected in the first stage, we compute attention between $\mathbf{q}_i$ and all tokens within these selected blocks. In the following sections, we will introduce the details about these two stages.

\subsubsection{Dynamic Contextual Block Selection}
The most critical component of InfLLM v2 is the relevance score computation between query tokens and key-value blocks. To avoid token-by-token relevance calculations, InfLLM~\citep{xiao2024infllm} selects representative tokens from each block to serve as the block representation, defining the relevance score as the dot product between the query token and these representative tokens. While this approach can capture important semantic information within blocks, the selection of representative tokens involves token-level computation and memory access, becoming one of InfLLM's efficiency bottlenecks. Therefore, we introduce fine-grained semantic kernels to capture block semantics and avoid token-level memory access. Besides, we require query heads in the same group to share the same key-value blocks to reduce memory access costs.

\textbf{Semantic Kernels}\quad InfLLM v2 further improves the relevance score computation method. Since sparse attention mechanisms need to minimize discontinuities in memory access, InfLLM v2 requires coarse-grained block partitioning of key-value sequences, with the block size $m$ typically being a relatively large value. If we use only one vector to represent the semantics of each block, we will inevitably encounter the problem of information loss.
To achieve more accurate relevance computation, InfLLM v2 introduces fine-grained \textit{semantic kernels} to construct representations for each key-value block.
Specifically, InfLLM v2 partitions the key-value sequence at a finer granularity, producing several semantic kernels. To ensure that each semantic span in the input sequence is contained within a complete semantic kernel, these kernels need to overlap with each other. Formally, InfLLM v2 divides the input sequence into semantic kernels of a size $p$, with a stride $s$ between adjacent kernels. Thus, $\mathbf{K}$ is partitioned into $\mathcal{S} = \{\mathbf{S}_0, \ldots, \mathbf{S}_{\lfloor\frac{l}{s}\rfloor-1}\}$, where $\mathbf{S}_{\hat{j}} = \mathbf{K}_{\hat{j}s:\hat{j}s+p}$. Since semantic kernels only participate in relevance computation, only the key vectors need to be retained.

InfLLM v2 uses the mean pooling operator to compute the representation of each semantic kernel $\text{Mean}(\mathbf{K}_{\hat{j}s:\hat{j}s+p})$, and the relevance score between a query token $\mathbf{q_i}$ and a semantic kernel is defined as $r_{\text{kernel}}(\mathbf{q_i}, \mathbf{S}_j) = \text{softmax}(\mathbf{q_i} \cdot \text{Mean}(\mathbf{K}_{\hat{j}s:\hat{j}s+p}))$. The relevance score between query token $\mathbf{q_i}$ and block $\mathbf{B}_j$ can then be represented as the maximum relevance score of all semantic kernels that intersect with $\mathbf{B}_j$:
\begin{equation}
r_{\text{block}}(\mathbf{q_i}, \mathbf{B}_j) = \max r_{\text{kernel}}(\mathbf{q_i}, \mathbf{S}_{\hat{j}}),\quad \mathbf{S}_{\hat{j}} \in \mathcal{S} \quad \text{and} \quad \mathbf{B}_j \cap \mathbf{S}_{\hat{j}} \neq \emptyset.
\end{equation}
Based on the relevance scores $r_{\text{block}}(\mathbf{q_i}, \mathbf{B}_j)$, InfLLM v2 selects the $k$ blocks with the highest relevance. Then, InfLLM v2 computes attention between the query token $\mathbf{q_i}$ and all tokens within these selected blocks to produce the final output $\mathbf{o_i}$. 

Notably, considering that the initial tokens as well as the tokens within the local window usually contribute a lot to the final outputs, InfLLM v2 sets the relevance scores between each query token $\mathbf{q_i}$ and the initial key-value blocks $\mathbf{B}_0$ and local window blocks to infinity. This mechanism ensures that each query token can attend both the initial blocks and the blocks within the local window. When the text length is short and does not exceed the total length of $k$ blocks, InfLLM v2 degrades to the vanilla dense attention mechanism.

\textbf{Top-K Blocks Sharing}\quad
Current LLM architectures typically adopt grouped query attention layers, where multiple queries share a single key-value head. In InfLLM v2, we require query heads within the same group to share the same top-k relevant blocks. This allows us to minimize memory access as much as possible. Specifically, after each query head computes relevance scores with semantic kernels, we average the relevance scores within the group and use this average as the relevance score for the query group.

\textbf{Efficient Top-K Implementation}\quad
Top-K selection involves three sequential steps: 1) Computing relevance scores between query tokens and each semantic kernel, followed by the softmax normalization. 2) Aggregating relevance scores for each semantic kernel across the query group dimension. 3) Selecting the top-K context blocks for each query token based on the aggregated scores. This operation constitutes the computational bottleneck in the sparse attention mechanism.

Traditional dense attention mechanisms typically employ the FlashAttention algorithm~\citep{flashattention} to reduce memory usage and accelerate attention computation. Specifically, FlashAttention leverages online softmax to minimize HBM access operations during the attention computation process. However, Top-K selection differs fundamentally from the attention computation process, as it requires precise numerical values of relevance scores between each query group and each semantic kernel. Since relevance computation, softmax normalization, and score aggregation across the query group dimension do not satisfy the commutative property, online softmax cannot be utilized. Consequently, Top-K selection requires two passes of memory access and computation on semantic kernels, where the first pass is used to compute the LogSumExp~(LSE) and the second pass is used to calculate the final relevance scores. 

Top-K selection is the bottleneck of InfLLM v2 for long-context processing. To reduce the computational costs, we propose an efficient LSE approximation method. Different from computing the dot product results between query tokens and all semantic kernels, we attempt to approximate the LSE value by introducing coarse-grained semantic kernels, whose kernel size $s_c$ is much larger than $s$. Then we compute the LSE of the relevance scores between query tokens and coarse-grained semantic kernels. The computational and memory access costs of this method require only $\frac{s}{s_c}$ of the original approach.

\subsubsection{Design Principles for Trainable Sparse Attention}
With the development of those applications requiring long-context processing and deep reasoning abilities, trainable sparse attention mechanisms show great potential to improve the efficiency of pre-training and inference.
In this section, we discuss several key features and design principles for InfLLM v2. We hope these discussions can promote future advancements of trainable sparse attention mechanisms. 

\textbf{Complexity Analysis}\quad
InfLLM v2 enables each token to compute attention with only the top-$k$ key-value blocks, significantly reducing the computational and memory access overhead of attention mechanisms. In this paragraph, we analyze the computational and memory access complexity of InfLLM v2. In stage 1, we need to calculate relevance scores between the query token and each semantic kernel. For a query token with the context length $l$, there are $\lfloor\frac{l}{s}\rfloor$ semantic kernels. Therefore, this query token requires $\lfloor\frac{l}{s}\rfloor$ vector multiplications and memory accesses. In stage 2, we need to compute the relevance scores between the query token and $k$ key-value blocks. During this process, the query token requires $2km$ vector multiplications and memory accesses. Compared to dense attention mechanisms, which require $2l$ vector operations and memory accesses, when the sequence is very long ($l\gg m$), InfLLM v2 can reduce computational overhead and memory accesses to $\frac{1}{s}$. We can see that the computational complexity of stage 1 remains $O(l^2)$, while the computational complexity of stage 2 is $O(l)$. Thus, if we want to further improve the efficiency of sparse attention, we are supposed to devote more efforts to relevant blocks retrieval. 

\textbf{Different Granularity for Query and Key-Value Tokens}\quad
In InfLLM v2, we allow each query token to compute attention with different key-value blocks. Here, the computational unit for queries is token-level, while the computational unit for key-values is block-level. Many previous sparse attention approaches~\citep{xu2025xattention,zhang2025spargeattn} also segment the query sequence into blocks, where all query tokens within a block share the same top-$k$ key-value blocks. However, query blocking operations can only accelerate long-sequence prefilling but cannot speed up the decoding process, as decoding requires token-by-token generation, and in most cases, query tokens cannot form a complete block. Therefore, using block-level units for queries during training leads to training-inference inconsistency during decoding, which subsequently degrades model performance.

\textbf{Trainable Context Selection}\quad
In sparse attention mechanisms, the top-$k$ selection operation is non-differentiable. This means that the representation of semantic kernels cannot be optimized through the sparse attention computation in stage 2. Therefore, in InfLLM v2, we choose to use mean pooling, a parameter-free operation, to construct the representation of semantic kernels. Additionally, mean pooling has the advantage of ensuring that the representation of semantic kernels remains in the same semantic space as token-level key vectors. In this way, we can indirectly optimize the representation of semantic kernels by optimizing token-level key vectors. NSA~\citep{yuan2025native} employs a compressed attention mechanism, adding the output of context selection to the attention output, thereby enabling optimization of block representations. However, this operation adds significant overhead for short texts. In contrast, the mean pooling operation adopted by InfLLM v2 does not affect efficiency for short texts, offering greater practical value.

\textbf{Hyper-Parameter Recommendation}\quad
Efficient training and inference of sparse attention mechanisms require highly coordinated algorithm design and operator design. Therefore, from the perspectives of algorithmic effectiveness and hardware constraints, the hyperparameter design in InfLLM v2 needs to satisfy certain conditions. Regarding the choice of semantic kernel size, smaller kernel sizes enable more precise relevance score computation; however, smaller kernel sizes also incur greater computational overhead. Thus, to achieve a good balance between effectiveness and efficiency, we chose to set the kernel size to 32 and the stride to 16. Constrained by the Matrix Multiplication Accumulation instructions of GPU tensor cores, a query group must contain at least 16 heads to ensure hardware is fully utilized.





\subsection{UltraClean: High-Quality Pre-Training Data Filtering and Generation}
With the rapid development of LLMs, data quality has become one of the key factors in improving model performance. Therefore, to enhance the capability density of \modelname{}, we conduct extensive data engineering, including filtering high-quality knowledge-intensive data from massive internet sources and utilizing existing LLMs to generate high-quality reasoning-intensive data.
Specifically, we introduce UltraClean, an efficient pre-training filtering technology, based on which we can achieve comparable performance with Qwen3-8B trained with 36 trillion tokens using only 8 trillion tokens. 

\subsubsection{High-Quality Knowledge-Intensive Data Filtering}
\begin{figure*}[t]
    \centering
    \includegraphics[width=0.98\textwidth]{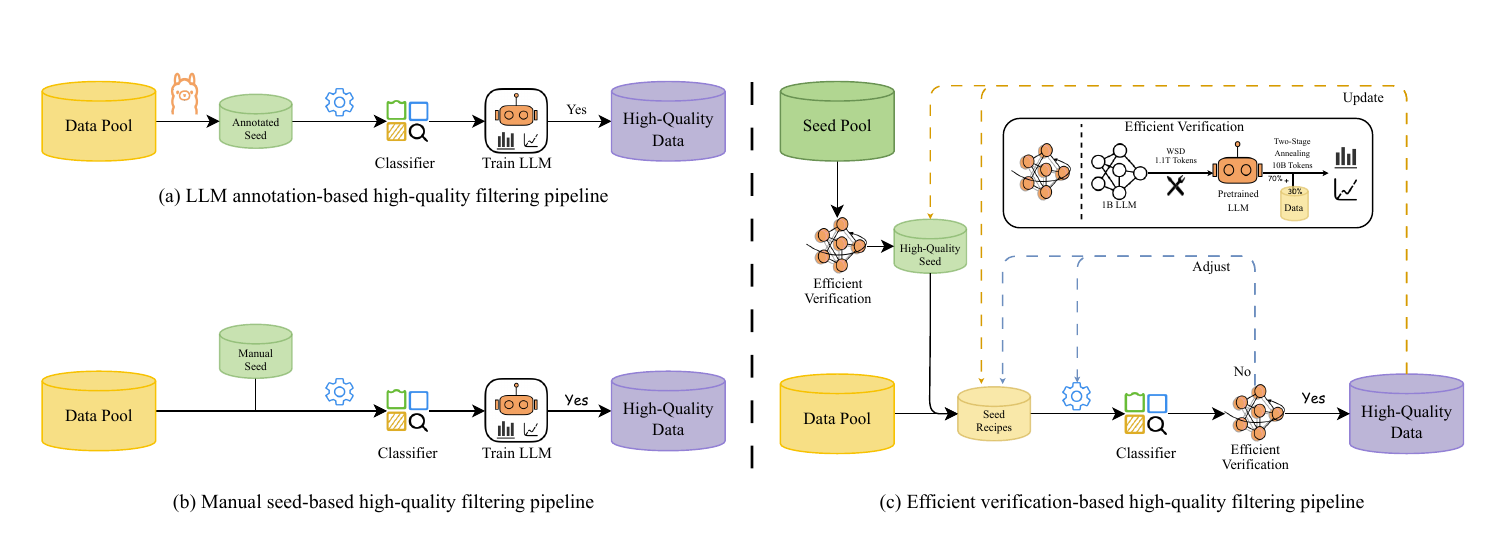}
    \caption{The illustration of high-quality data filtering pipelines. Traditional model-based data filtering methods (a) and (b) rely on human expertise for seed data selection and lack data quality verification.}
    \label{fig:ultrafineweb}
\end{figure*}

With the rapid development of LLMs, data quality has become one of the key factors for improving model performance. Leveraging model-based classifiers to filter data and extract high-quality knowledge-intensive samples can not only enhance model effectiveness but also reduce training costs by achieving better performance with fewer tokens. However, current approaches still face two major challenges: (1) the lack of efficient data verification strategies makes it difficult to provide timely feedback on data quality; (2) the selection of seed data for training classifiers lacks clear standards and heavily relies on human expertise, which introduces subjective biases. To address these issues, we propose an efficient data verification strategy that can rapidly evaluate the actual impact of data on LLM training at minimal computational costs. Building on this, we optimize the positive and negative sample selection process based on the hypothesis that high-quality seed data contributes to better LLM performance, and construct an efficient data filtering pipeline. Our approach yields a more robust and higher-quality knowledge-intensive classifier, while also significantly improving the overall quality of pre-training data.

\textbf{Overall Workflow}\quad
The overall workflow is illustrated in Figure~\ref{fig:ultrafineweb}(c). We begin by applying the efficient verification strategy to evaluate the initial pool of candidate seed samples, selecting high-quality data that significantly improves training performance as positive seeds for classifier training. Meanwhile, negative samples are randomly drawn from the raw data pool to construct a balanced training set. To more efficiently assess the actual effectiveness of the classifier, we also apply an efficient verification strategy to evaluate its filtering results. Based on the feedback from verification, we iteratively update the high-quality seed pool, dynamically adjust the ratio of positive to negative samples, and fine-tune the training hyperparameters of the classifier, thereby continuously optimizing the data filtering strategy. Only classifiers demonstrating stable and reliable performance under efficient verification are used for large-scale data filtering and subsequent model training. It is worth emphasizing that the final high-quality classifier is applied to the entire web-scale pre-training dataset to extract high-quality knowledge-intensive training samples, fundamentally improving the efficiency and effectiveness of large-scale model training.

\textbf{Efficient Verification Strategy}\quad
Under limited token budgets, performance differences in LLM training often fail to reach statistical significance, and the inherent instability of the training process further undermines the reliability of verification results. Effective verification of pre-training data typically requires at least 100B tokens. As shown in Table~\ref{tab:comparison_train_cost}, training 100B tokens on an LLM with 1B parameters requires approximately 1,200 GPU hours, equivalent to running 64 GPUs continuously for nearly 19 hours. Such high computational costs make it impractical to perform efficient verification during the iterative development of high-quality data classifiers. To address this, we draw inspiration from the design of Llama 3.1~\citep{llama3.1} and propose an efficient verification strategy. Specifically, we pre-train a 1B-parameter LLM using the WSD scheduler~\citep{minicpm}, covering a total of 1.1 trillion (T) tokens. This includes a stable training phase over 1T tokens and a decay phase over an additional 0.1T tokens. On this foundation, we introduce a two-stage annealing training process, where fine-tuning is conducted on 10B tokens, with 30\% of the data reserved for validation and the remaining 70\% following the default mixed-data ratio. Compared to the full training cost of 1,200 GPU hours, this strategy reduces training time to approximately 110 hours (i.e., fewer than 3.5 hours on 32 GPUs), significantly lowering computational demands and greatly improving the efficiency and iterability of the data filtering pipeline. We use a two-stage annealing training process with the original mixed-data ratio as a comparison baseline. This verification strategy enables efficient evaluation of the impact of candidate data on model training across multiple metrics, balancing accuracy and cost-effectiveness, and providing a practical and closed-loop optimization path for the data selection process.

\begin{table}[t]
    \centering
    \small
    \caption{The comparison of the computational costs across different verification strategies on an LLM with 1B parameters.}
    \renewcommand{\arraystretch}{1.2}
    \begin{tabular}{lccc}
        \toprule
         & 100B from scratch &  380B from scratch & Efficient verification Strategy  \\
        \midrule
        GPU Hours &  1,200          &  4,600  &  \textbf{110} \\
        \bottomrule
    \end{tabular}
    \label{tab:comparison_train_cost}
    \vspace{-1em}
\end{table}

\textbf{Classifier Training Recipes}\quad
Currently, the selection of positive seed samples for classifier training primarily relies on LLM scoring or manual curation. However, the former is susceptible to biases inherent in LLMs, which may introduce systematic noise and annotation artifacts, while the latter depends heavily on human expertise and lacks rigorous evaluation of seed data effectiveness. Moreover, the reliability of manual selection is often judged indirectly through the downstream performance of LLMs trained on the filtered data, making the verification process both expensive and indirect. These limitations hinder the adaptability and generalization ability of classifiers across different tasks. To address this, we propose a core hypothesis: high-quality seed data that can improve LLM performance should also be beneficial for training classifiers capable of identifying high-quality training samples. As shown in Figure~\ref{fig:ultrafineweb}(c), we first apply our efficient verification strategy to the initial pool of candidate seed samples and select those that yield significant performance gains in LLM training as positive samples for classifier training. We evaluate a large number of candidate seeds and ultimately curate a set of high-quality, empirically verified positive samples. These include: high-quality in-domain data with LLM scores above 4, instruction-formatted datasets (\textit{e.g.}, OH-2.5 and ELI5), real-world educational materials, LLM-synthesized textbook-style content, and curated high-quality web data. This process not only ensures the overall quality of positive samples but also significantly improves filtering efficiency, providing a strong foundation for classifier training. To enhance classifier robustness, we construct negative samples using raw data from diverse sources. English negative samples are drawn from FineWeb, C4, Dolma, The Pile, and RedPajama, while Chinese negatives include CCI3, ChineseWebtext, and other mainstream corpora. Experimental results further demonstrate that incorporating diversified data sources for negative samples significantly improves classifier generalization and cross-domain adaptability. After the initial training, we adopt an iterative training mechanism: the positive and negative samples inferred by the current classifier version are used as new training data for the next round, continuously optimizing classifier performance. Through this iterative process, we further improve the precision and stability of the classifier for large-scale data filtering tasks.

\textbf{FastText-based Quality Filtering}\quad
Current high-quality data classifiers can be broadly categorized into LLM-based approaches~\citep{penedo2024fineweb, chinese_fineweb} and fastText-based methods~\citep{dclm, deepseekmath}. While LLM-based classifiers generally offer strong performance, they incur significantly higher inference costs. To address this, we adopt a fastText-based classifier, which drastically reduces inference overhead while maintaining competitive performance under certain conditions. This approach not only minimizes resource consumption but also accelerates the iteration cycle of data filtering experiments. For example, processing 15 trillion (15T) tokens using an LLM-based classifier requires approximately 6,000 hours on GPUs, whereas fastText can complete the same task on a non-GPU server using 80 CPUs in under 1,000 hours, offering a substantial efficiency gain. Notably, most of our large-scale experiments are conducted on a distributed Spark cluster~\footnote{\url{https://spark.apache.org/}}. In the data preprocessing stage, we apply a series of essential steps, including the removal of redundant blank lines and excessive whitespace, stripping of diacritics, and normalization of English text to lowercase. We use the tokenizer from DeepSeek-V2~\citep{deepseekv2}, which outperforms traditional tokenization methods (e.g., space-based tokenization for English and Jieba for Chinese~\footnote{\url{https://pypi.org/project/jieba/}}). At the same time, we preserve structural tokens such as \texttt{\textbackslash n}, \texttt{\textbackslash t}, and \texttt{\textbackslash r}. For training, we configure the fastText classifier with the following hyperparameters: the vector dimension is set to 256, the learning rate is set to 0.1, the maximum n-gram length is set to 3, the minimum word frequency is set to 5, and the number of training epochs is set to 3. During inference, we adopt the default classification threshold of 0.5 to simplify the workflow and ensure consistency across experiments, avoiding the need for additional hyperparameter tuning. During inference, we adopt a default classification threshold of 0.5 to simplify the workflow and ensure consistency across experiments, avoiding additional hyperparameter tuning.

\textbf{Results and Analysis}\quad
In our experiments, we utilize the MiniCPM-1.2B model architecture with the MiniCPM3-4B tokenizer. Each experiment involves training on approximately 100B tokens. 
We employ the Lighteval library~\citep{lighteval} for model evaluation, mirroring the setup used with FineWeb~\citep{penedo2024fineweb} and CCI3-HQ~\citep{cci3}. All evaluation metrics are based on a zero-shot setting.
As shown in Table~\ref{tab:ultra_combined_result}, on the English metrics, UltraFineWeb-en demonstrates significant improvements in performance on multiple tasks, including MMLU, ARC-C, ARC-E, CommonSenseQA, and OpenBookQA. 
Specifically, UltraFineWeb outperforms FineWeb in these tasks, with only a slight drop of 0.15 percentage points ($pp$) in HellaSwag compared to FineWeb, but a 0.6$pp$ improvement over FineWeb-edu. 
The English average score for UltraFineWeb-en (45.891$pp$) is 3.61$pp$ higher than that of FineWeb (42.287$pp$) and 1.3$pp$ higher than FineWeb-edu (44.560$pp$). 
On the Chinese metrics, UltraFineWeb-zh also outperforms both FineWeb-zh and FineWeb-edu-zh on C-Eval and CMMLU. 
Specifically, UltraFineWeb-zh improves by 0.31$pp$ and 3.65$pp$ over Chinese FineWeb and Chinese FineWeb-edu-v2 on C-Eval and CMMLU, respectively, and by 0.09$pp$ and 0.13$pp$ compared to FineWeb-edu-zh. 
The Chinese average score for UltraFineWeb-zh increases by 1.98$pp$ and 0.61$pp$, respectively, compared to FineWeb-zh and FineWeb-edu-zh. These results indicate that our proposed High-Quality Data Filtering Pipeline significantly improves data quality, leading to notable improvements in model performance.

\begin{table*}[t]
    \centering
    \small
    \caption{The comparison of individual results on English and Chinese datasets.}
    \renewcommand{\arraystretch}{1.2}
    \begin{tabular}{llll}
        \toprule
        Metrics                         & FineWeb           & FineWeb-edu                           & UltraFineWeb-en \\
        \midrule
        MMLU                          &  28.84          &  31.80$_{\textcolor{red}{+2.96}}$     &  \textbf{32.24}$_{\textcolor{red}{+3.40}}$    \\
        ARC-C                         &  25.17          &  34.56$_{\textcolor{red}{+9.39}}$     &  \textbf{35.67}$_{\textcolor{red}{+10.50}}$   \\
        ARC-E                         &  59.18          &  69.95$_{\textcolor{red}{+10.77}}$    &  \textbf{70.62}$_{\textcolor{red}{+11.44}}$  \\
        CommonSenseQA                 &  34.32          &  31.53$_{\textcolor{blue}{-2.79}}$    &  \textbf{36.45}$_{\textcolor{red}{+2.13}}$   \\
        HellaSwag                     &  \textbf{42.91}   &  42.17$_{\textcolor{blue}{-0.74}}$    &  42.76$_{\textcolor{blue}{-0.15}}$  \\
        OpenbookQA                      &  22.20          &  25.20$_{\textcolor{red}{+3.00}}$     &  \textbf{26.20}$_{\textcolor{red}{+4.00}}$   \\
        PIQA                          &  73.29          &  72.14$_{\textcolor{blue}{-1.15}}$  &  \textbf{73.67}$_{\textcolor{red}{+0.38}}$   \\
        SIQA                          &  38.95          &  38.13$_{\textcolor{blue}{-0.82}}$  &  \textbf{39.61}$_{\textcolor{red}{+0.66}}$   \\
        Winogrande                      &  55.64          &  55.56$_{\textcolor{blue}{-0.08}}$  &  \textbf{55.80}$_{\textcolor{red}{+0.16}}$   \\
        \midrule
        Average    &  42.28          &  44.56$_{\textcolor{red}{+2.282}}$ &  \textbf{45.89}$_{\textcolor{red}{+3.613}}$ \\
        \midrule
        \midrule
        Metrics                             &  Chinese-FineWeb  & Chinese-FineWeb-edu-v2            & UltraFineWeb-zh \\
        \midrule
        C-Eval                              &  33.95          & 34.17$_{\textcolor{red}{+0.22}}$    & \textbf{34.26}$_{\textcolor{red}{+0.31}}$ \\
        CMMLU                             &  32.41          & 34.93$_{\textcolor{red}{+2.52}}$     & \textbf{36.06}$_{\textcolor{red}{+3.65}}$ \\
        \midrule
        Average       &  33.18          & 34.55$_{\textcolor{red}{+1.37}}$    & \textbf{35.16}$_{\textcolor{red}{+1.98}}$ \\
        \bottomrule
    \end{tabular}
    \label{tab:ultra_combined_result}
    \vspace{-1em}
\end{table*}

\subsubsection{High-Quality Reasoning-Intensive Data Generation}
In the pre-training process of building high-performance LLMs, reasoning ability is widely considered to be one of the core indicators of general intelligence. The development of this ability largely depends on the quality, structure, and knowledge density of pre-training data.
However, current typical pre-training corpora, such as Common Crawl and large-scale web-crawled datasets, still face significant challenges in supporting complex reasoning skills. Specifically, two core issues are commonly observed in existing corpora: (1) Although web data features diverse linguistic forms and broad coverage, much of it consists of advertisements, template texts, superficial dialogues, and various redundant information. These contents typically exhibit low knowledge density and weak logical structure. Even with high-quality data classifiers for filtering, the extracted content often lacks accurate reasoning chains and systematic knowledge organization, making it difficult for models to acquire transferable thought patterns and reasoning paradigms. (2) Current data construction processes have a clear scale-depth trade-off. The corpora for training LLMs heavily rely on long-tail content with low information density to pursue broad coverage and diversity. In contrast, truly structured and knowledge-rich texts, such as textbooks, academic materials, and systematically explanatory content, constitute only a small fraction. This imbalanced data structure limits the model’s ability to learn deep semantic relationships and complex logical structures. As a result, while models may generate fluent language, they often lack depth in reasoning and may even produce hallucinations in reasoning-intensive tasks.

To address these challenges, we propose a high-quality, reasoning-intensive data generation pipeline to improve reasoning capabilities. This task-oriented and training-efficient pipeline integrates seed data selection with structured data curation and generation. Through multi-round iteration, we gradually build pretraining data that is high in knowledge density, clearly structured, logically coherent, and linguistically diverse. This not only improves model performance on general benchmarks, but also fosters stronger reasoning transferability and generalization at a foundational capability level.

\textbf{Seed Data Selection}\quad
For selecting seed data, we focus on two core objectives: high knowledge density and domain diversity. On the one hand, we leverage the high-quality data classifier built on the UltraFineWeb dataset to filter out knowledge-intensive, logically complete, and well-structured content from large-scale web corpora. The selected content can serve as high-quality seed data for general domains.
These data fragments exhibit strong contextual coherence and conceptual richness, offering structured linguistic templates and knowledge expressions for synthetic models. On the other hand, to support higher-order domain-specific reasoning capabilities, we conduct manual curation targeting key fields such as mathematics, programming, and natural sciences. We select high-quality open-source web content, textbooks, and QA materials as domain-specific seeds. These resources typically feature highly structured expressions and clear knowledge organization. By combining automated filtering with human curation, we construct a seed pool that balances generality and specialization, laying a solid foundation for subsequent reasoning-intensive data generation.

\textbf{Structured Data Curation and Generation}\quad
To effectively enhance the knowledge density and reasoning capabilities, we design a structured data curation and generation mechanism centered on clear structure, semantic richness, and logical progression. Traditional web corpora often present information in fragmented and unstructured forms. To address this, we adopt a structure-first principle, using open-source LLMs under 10B parameters to automatically edit and restructure web text. Key operations include denoising and cleansing, semantic integration, and logical completion, resulting in mid-length to short-length passages that are well-formed, logically coherent, and hierarchically organized.

Additionally, based on high-quality open-source web corpora, textbooks, and QA materials, we abstract two typical paradigms of knowledge construction (textbook and forum), and develop corresponding synthetic strategies. For the textbook paradigm, we emphasize systematicity and progression, building a layered structure consisting of knowledge points, multi-round explanations, summaries, and practice questions. For the forum paradigm, we simulate authentic user discussions by generating multi-turn QA exchanges and viewpoint debates around a central topic, thereby enhancing diversity and realism. These generation approaches not only preserve high knowledge density but also offer multiple reasoning paths and cognitive frameworks for the model, facilitating the development of more generalized and critical-thinking language models.

Notably, the constructed data are fed back into the original seed pool for subsequent rounds of evolutionary generation. This gives the data generation process a self-enhancing nature, whereby multi-round iterative evolution progressively builds richer and deeper high-quality reasoning corpora, providing the model with continuously optimized training signals and a more resilient cognitive support structure.

\subsubsection{Discussion for Future Training Data}

We systematically introduce the high-quality data construction strategies employed in the pretraining of \modelname{}, covering high-quality knowledge-intensive data filtering and reasoning-intensive data generation. Without increasing, and in some cases even reducing, the training tokens, we significantly improve the knowledge density and logical complexity of the corpus through automated filtering and structured generation. The experimental results show that this strategy yields performance comparable to, or even surpassing, that of models trained on the full-scale corpus across a range of downstream tasks. In particular, for knowledge-intensive and reasoning-intensive tasks, the optimized data more effectively enhances models' general capabilities and knowledge transfer abilities, further validating the principle that data quality outweighs data quantity.

Despite the promising results, several open questions remain for future exploration. First, the current data filtering and generation pipelines still rely partially on human-designed heuristics. Leveraging the capabilities of LLMs to construct self-supervised mechanisms for data quality assessment and structural optimization is a key direction for improving both efficiency and quality. Second, in the evolutionary generation process, balancing corpus diversity with task relevance remains challenging, requiring finer-grained feedback mechanisms to prevent semantic mode collapse. Lastly, extending this strategy to multilingual, cross-task, and even multimodal settings represents a crucial step toward building the next generation of high-performance foundation models.

\subsection{ModelTunnel v2: Efficient Pre-Training Strategy Search}
Training LLMs requires enormous computational costs, making it a critical challenge to maximize model performance while minimizing computational resource consumption. In our previous work~\citep{minicpm}, we have built a ModelTunnel based on predictable scaling technology. This enables us to search for training strategies on small models and transfer them to train large models, thereby reducing the experimental costs of determining optimal training configurations for large models.
During the training process of \modelname{}, we reuse the relevant configurations from our ModelTunnel and develop ModelTunnel v2, which features improvements for search precision and provides systematic validation of $\mu$P's effectiveness. Furthermore, to enhance model training efficiency, we implement engineering optimizations in both the pre-training objectives and the infrastructure. Next, we will provide detailed descriptions of these improvements for the pre-training of \modelname{}.

\subsubsection{Efficient Predictable Scaling with Improved Performance Indicator}

As model parameter scales continue to increase, the cost of conducting model training experiments rises correspondingly, with a single model training requiring hundreds of thousands of GPU hours. This makes traditional methods like grid search for model training configuration search increasingly impractical. In MiniCPM-1, we systematically construct a ModelTunnel that conducts extensive experiments on models with only millions of parameters to determine optimal hyperparameters. In constructing \modelname{}, we make improvements in the following aspects:

1) \textbf{More Reasonable Performance Indicators}\quad
In MiniCPM-1, we use the model's language model loss on open-source pretraining corpora as the performance indicator, assuming that lower language model loss indicates superior model performance. However, loss on open-source pretraining datasets cannot accurately reflect model performance on downstream tasks. Therefore, we construct ScalingBench and establish the relationship between loss on the ScalingBench and downstream task performance~\citep{xiao2024densing}.

Since most models trained in the Wind Tunnel have very limited parameters and training data, they often fail to demonstrate non-random performance on downstream tasks, making direct evaluation of small models from the Wind Tunnel on downstream tasks unreliable. Therefore, we attempt to construct an evaluation dataset where the loss correlates well with downstream task performance through a functional mapping relationship. To achieve this goal, we construct ScalingBench from the validation datasets of downstream tasks. In the original downstream datasets, each instance consists of a user instruction and a human-annotated label that usually contains a few words.  
In ScalingBench, we use GPT-4o~\citep{gpt4} to generate reasoning steps for all test instances. Then we directly calculate the conditional loss on the reasoning steps and labels, specifically the loss incurred when the model generates answers given task inputs. The loss can serve as a reasonable performance indicator.

\begin{figure*}[t]
    \centering
    \includegraphics[width=0.98\textwidth]{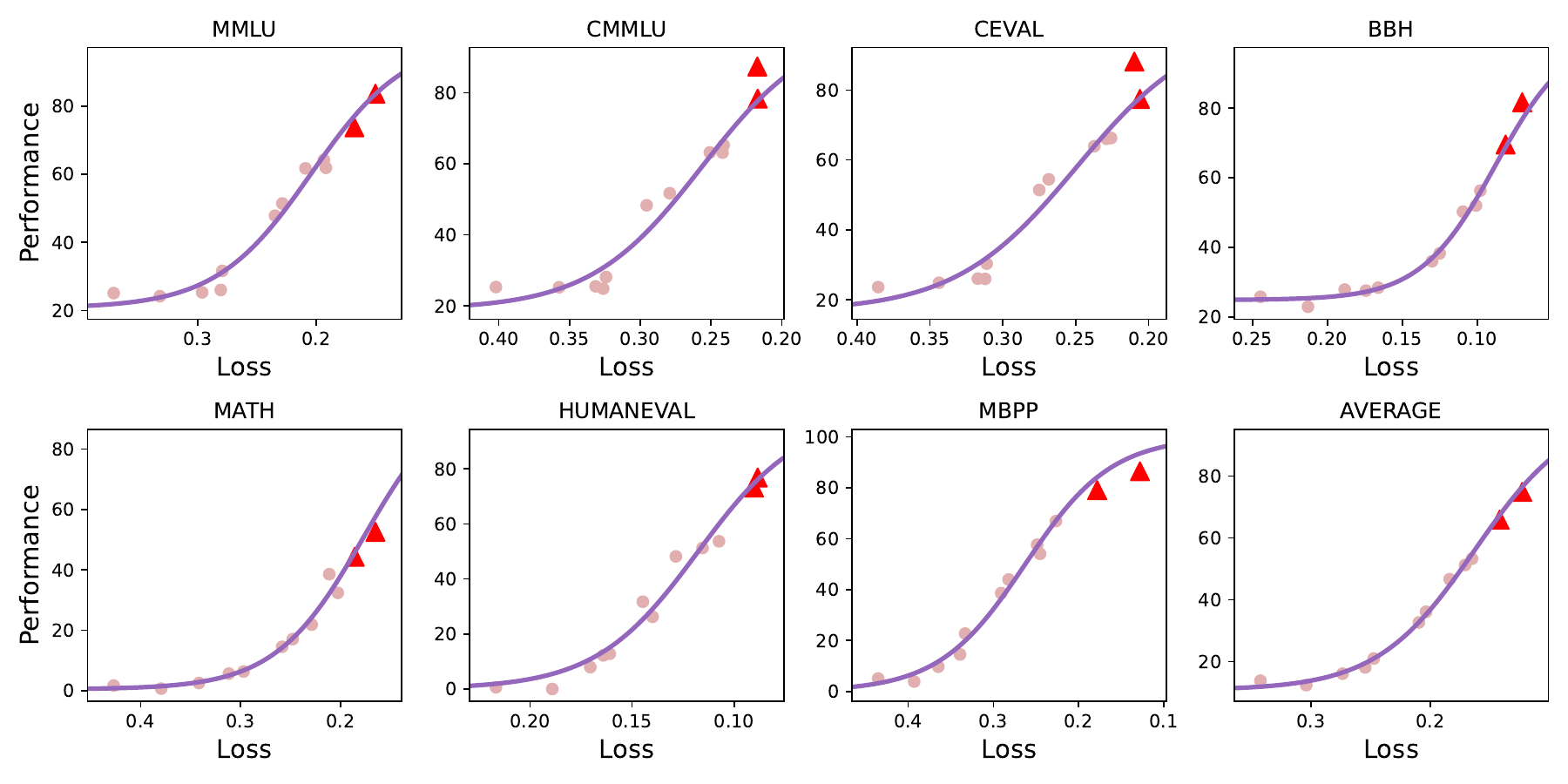}
    \caption{The sigmoid relationship between loss and downstream performance on ScalingBench.}
    \label{fig:scalingbench}
    \vspace{-1em}
\end{figure*}

To verify the effectiveness of ScalingBench and establish the relationship between the loss and downstream performance, we evaluate the loss and performance of multiple models on ScalingBench. These models are trained by our team at different periods with identical vocabularies, making their losses comparable. The models range from 0.36B to 4B parameters with varying training data sources and scales. All models demonstrate the same trend on ScalingBench: a sigmoid function relationship between ScalingBench loss and downstream performance. As shown in Figure~\ref{fig:scalingbench}, red triangles represent 7B and 80B parameter models that do not participate in function fitting and serve as test points for the curve. Their ScalingBench scores and performance on the corresponding tasks are also consistent with the sigmoid relationship. These results show that our constructed ScalingBench effectively establishes the relationship between loss and downstream task performance, demonstrating that loss on ScalingBench serves as a highly effective performance indicator.
2) \textbf{Comparison between $\mu$P and Vanilla Architecture}\quad
Leveraging predictable scaling for hyperparameter search represents a critical pathway for reducing experimental costs while maximizing model performance. This direction has garnered significant attention recently, and research in this area broadly falls into two categories: architecture-driven hyperparameter transfer~\citep{yang2022tensor,everett2024scaling,lingle2024large,bordelondepthwise} and data-driven hyperparameter transfer~\citep{scaling-law,bjorck2024scaling,li2025predictablescalei}. The former modifies the model's computational process to ensure hyperparameter settings can be shared between small and large models. The latter determines optimal hyperparameter configurations for LLMs by analyzing the relationship between hyperparameters and model parameter scale. In MiniCPM-series models, we adopt $\mu$P~\citep{yang2022tensor} as our basic architecture, which assumes that hyperparameters such as learning rate can be transferred between different model sizes under certain constraints. And most existing open-source models adopt the data-driven methods to search for hyperparameters.

In \modelname{}, we compare the $\mu$P architecture with the vanilla model format. Recently, various approaches to hyperparameter search have been proposed, with StepLaw~\citep{li2025predictablescalei} being one notable method in this area. However, our analysis reveals a significant discrepancy between the experimental configurations in the original StepLaw study and our practical training requirements. Given the potential benefits of both StepLaw and $\mu$P for hyperparameter optimization, we seek to systematically evaluate their relative performance in our specific training context. To this end, we design a comprehensive comparative study between StepLaw and the $\mu$P method.

We conduct our experiments by first computing predicted hyperparameters using StepLaw, specifically focusing on batch size and learning rate predictions. We then implement batch size selection through nearest-neighbor quantization by choosing the 16-divisible value closest to the StepLaw batch prediction to ensure computational efficiency. We compare two distinct training configurations: a standard architecture without $\mu$P using the learning rate predicted by StepLaw, and a $\mu$P architecture with empirically optimized learning rate parameters. We evaluate the performance of both trained models using training loss metrics and ScalingBench, providing a comprehensive assessment of each approach's effectiveness in our specific training context.

As shown in Table~\ref{tab:steplaw}, steplaw demonstrates slightly more instances of advantage, but the differences in both loss values and scalingbench scores between the two methods remain minimal, with neither approach exhibiting consistently stable superiority. We attribute the comparable effectiveness of hyperparameter search in the $\mu$P framework to steplaw under our experimental conditions to the following factors: 1)~Practical hardware constraints prevent strict adherence to steplaw's prescribed batch size. 2)~The experiments employ the WSD learning rate schedule, and different data allocations are used for stable and decay phases, whereas step-law experiments utilize cosine decay. 3)~We apply ScalingBench in the model evaluation, while step-law's fitting process relies on training loss. Additionally, randomness exists in both the training process and evaluation metrics. We further note that reproducing steplaw's work incurs significant expense, whereas $\mu$P search requires minimal GPU hours. This makes it accessible to common researchers. In conclusion, we believe that $\mu$P with a hyperparameter search method can use real training configurations and data to conduct a low-cost hyperparameter search in a new experiment. Steplaw still shows strong performance in experimental environments with many changes, and in many cases, can be used as a comparison standard for parameter search methods.

\begin{table}[t]
\caption{Comparison of hyperparameter search computational cost and performance between $\mu$P and StepLaw.}
\small
\centering
\begin{tabular}{ll|c|cccccccc}
\toprule
\multicolumn{2}{l|}{\# Model Params} & \multirow{2}{*}{\makecell[c]{GPU\\Hour}}& 150M & 150M & 150M & 360M & 360M & 360M & 700M & 700M \\
\multicolumn{2}{l|}{\# Training Tokens} & & 4B & 10B & 20B & 20B & 40B & 100B & 40B & 100B \\ \midrule
\multirow{2}{*}{LM Loss} 
    & Vanilla & 1M & 2.1834 & \textbf{2.0335} & \textbf{1.9608} & \textbf{1.8510} & 1.7755 & 1.7274 & \textbf{1.6956} & \textbf{1.6391} \\
    & $\mu$P & 32 & \textbf{2.1797} & 2.0491 & 1.9666 & 1.8547 & \textbf{1.7721} & \textbf{1.7174} & 1.7091 & 1.6419 \\ \midrule
\multirow{2}{*}{ScalingBench} 
    & Vanilla & 1M & \textbf{0.4310} & \textbf{0.3897} & 0.3685 & 0.3383 & \textbf{0.3172} & \textbf{0.3004} & \textbf{0.2984} & 0.2805 \\
    & $\mu$P & 32 & 0.4434 & 0.3915 & \textbf{0.3677} & \textbf{0.3371} & 0.3196 & 0.3045 & 0.3013 & \textbf{0.2789} \\  
 \bottomrule
\end{tabular}
\label{tab:steplaw}
\end{table}



\subsubsection{Pre-Training Engineering}
To improve model training efficiency, inspired by \cite{deepseek-v3}, we adopt multi-token prediction~\citep{gloeckle2024better} as our training objective to introduce denser supervision signals and improve data efficiency. For training infrastructure, we implement an FP8 mixed-precision computation framework.

\textbf{Multi-Token Prediction}\quad
Traditional LLM pre-training usually adopts next token prediction as the training objective, which requires the model to predict the next token based on the preceding context. Multi-token prediction (MTP) requires the LLM to output multiple tokens with additional prediction heads. Here, the additional prediction heads are a one-layer Transformer~\citep{transformer} with embedding layers and output heads shared with the main model. Specifically, given the input tokens $\{x_0, x_1, ..., x_{l-1}\}$, the main model will generate a sequence of hidden vectors $\mathbf{H} = \{\mathbf{h}_0, \mathbf{h}_1, ..., \mathbf{h}_{l-1}\}$. Then the next token prediction objective can be computed as follows:
\begin{equation}
    \mathcal{L}_{\text{NTP}} = \frac{1}{l}\sum_{i}\text{CrossEntropy}\left(\text{OutputHead}(\mathbf{h}_i), x_{i+1}\right).
\end{equation}
Then the inputs of the additional prediction heads are the concatenation of hidden vectors and the embedding of the next tokens:
\begin{equation}
    \mathbf{\hat{h}}_i = \text{Concat}(\text{Norm}(\mathbf{h}_i), \text{Norm}(\text{Emb}(x_{i+1}))).
\end{equation}
Then the MTP training objective can be defined as:
\begin{align}
    \{\mathbf{h}^{\text{MTP}}_0, \mathbf{h}^{\text{MTP}}_1, ..., \mathbf{h}^{\text{MTP}}_{l-2}\} &= \text{Transformer}(\text{Linear}(\{\mathbf{\hat{h}}_0, \mathbf{\hat{h}}_1, ..., \mathbf{\hat{h}}_{l-2}\})), \\
    \mathcal{L}_{\text{MTP}} & = \frac{1}{l-1}\sum_{i}\text{CrossEntropy}\left(\text{OutputHead}(\mathbf{h}^{\text{MTP}}_{i}, x_{i+2})\right).
\end{align}
The final pre-training objective is the weighted sum of these two training objectives: $\mathcal{L} = \mathcal{L}_{\text{NTP}} + \lambda \mathcal{L}_{\text{MTP}}$.

\textbf{FP8 Mix-Precision Training}\quad
Considering that NVIDIA's Tensor Core GPUs have powerful FP8 computing capabilities, we adopt the FP8
mix-precision in training. Following \cite{deepseek-v3},
we apply online block-wise FP8 quantization to both parameters and activations, using a block size of 128×128 for parameters and 128×1 for activations.
After quantization, we adopt the FP8 Matrix Multiply Accumulate (MMA) instruction and use FP32 as the accumulator's precision. To avoid the instability caused by FP8, we only apply FP8 on the linear projection. Specifically, we employ FP8 only in the computation of activations during the forward pass and in the calculation of input gradients during the backward pass. 
Considering that parameters are extremely sensitive to precision, the parameter gradients are computed in BF16 format.



\section{Efficient Post-Training}  
In this section, we present our approach to post-training large models after the pre-training phase, enabling them to leverage the world knowledge acquired during pre-training to follow user instructions. During the supervised fine-tuning stage, we construct a diverse and comprehensive set of instructions to fully activate the capabilities of the LLMs, ensuring a strong foundational competence. To this end, following our previous work~\citep{ding2023enhancing}, we construct UltraChat v2, a large-scale SFT dataset covering abilities including knowledge application, reasoning, tool use, and long context processing. 

Furthermore, to enhance the model's deep reasoning ability, we employ supervised fine-tuning augmented with long chain-of-thought reasoning and integrate reinforcement learning techniques. The rollout process of RL usually suffers from the unbalanced load challenge, which can lead to very inefficient computations. To address this issue, we propose a load-balanced RL strategy -- chunk-wise rollout. 

In the following paragraphs, we will provide details on supervised fine-tuning and reinforcement learning techniques used to enhance foundational capabilities and reasoning abilities.

\subsection{UltraChat v2: Foundational Capability Enhanced SFT Data Generation}
To improve LLMs' core competencies, we design a synthetic data generation framework focused on task-oriented capabilities. Guided by key capability dimensions, the framework generates high-quality QA-style data covering a wide range of skills, providing more targeted and structured signals during post-training.
We systematically develop synthetic data tracks across five key skill areas: knowledge applications, reasoning, instruction following, long-context processing, and tool use. Each track is tailored to the input-output patterns and cognitive demands of its target skill, providing diverse, task-driven, and transferable training examples.
This approach helps the model improve consistently across key foundational capabilities.

\subsubsection{Knowledge-Intensive Data}
We begin by extracting and organizing knowledge points from domain-specific corpora, exam syllabi, and textbook materials across various disciplines, thereby constructing a comprehensive and well-structured knowledge framework. Based on this framework, we leverage LLMs to generate practice question-answer pairs targeting individual knowledge points, thus forming the initial stage of a knowledge-driven QA dataset.

To further enhance the diversity and generalization capability of the data, we apply two evolution strategies to the initial practice QA pairs: (1) \textbf{Instruction evolution}, which rewrites prompts in diverse ways to simulate various questioning styles and task formulations; and (2) \textbf{Answer diversity evolution}, which guides the model to generate plausible and stylistically varied answers, thereby improving the model’s robustness in knowledge understanding and expression.

\subsubsection{Reasoning-Intensive Data}
Reasoning ability is a fundamental skill that enables LLMs to manage complex tasks and generalize knowledge across diverse domains. Unlike standard question-answering tasks, reasoning tasks demand more than factual retrieval. They require models to perform multi-step logical inference and integrate information across contexts, while maintaining coherence and structural clarity in their responses. To improve LLMs' abilities in mathematical reasoning, logical modeling, and procedural thinking, we develop two specialized datasets: one focused on mathematical reasoning, and the other on code-based reasoning. These datasets are designed to strengthen the model’s reasoning capacity and support the development of more robust and transferable logical skills.

\textbf{Math Reasoning Data}\quad
We start by systematically categorizing mathematical knowledge across core domains, including linear algebra, calculus, probability, statistics, differential equations, discrete mathematics, and differential geometry.
These topics are further organized by educational level---from elementary to university---to establish a clear hierarchy of difficulty.
Based on these topics, we either use curated seed data or prompt LLMs directly to generate relevant questions. The model is then guided to produce both answers and self-reflections to improve logical consistency and correctness.

In our methodology for generating math problems, we focus on two dimensions: instruction diversity and solution path variety. 
On the one hand, we extend basic problems into various formats---including multiple choice, fill-in-the-blank, and open-ended questions---to enhance both expressive diversity and structural complexity.
On the other hand, we generate multiple valid solution paths for the same problem, thereby expanding the model’s ability to explore the reasoning space and generalize its problem-solving strategies. The generation process is guided by a set of heuristic rules, such as instruction adherence and response length control, to ensure mathematical validity and answer verifiability. This results in a math-focused data subset with both instructional value and reasoning depth. 
Additionally, we implement difficulty-based stratification, actively reducing the proportion of easy questions during training. This encourages the model to focus on medium-to-high-difficulty problems, effectively strengthening its ability to construct reasoning chains and handle complex logical structures.

\textbf{Code Reasoning Data}\quad
To enhance the code programming capabilities of LLMs, we design a code reasoning data generation pipeline tailored to real-world development scenarios. First, we manually define various coding contexts, problem categories (such as semantic completion, bug localization, and complex logic understanding), and difficulty levels to ensure that the constructed tasks closely resemble real-world engineering scenarios and have explicit reasoning objectives. In the data construction process, we extract high-quality code snippets, such as function definitions, algorithmic segments, and class structures, from authentic GitHub repositories, coding challenge libraries, or open-source scripts to serve as contextual foundations for question generation. Leveraging these predefined contexts and snippet structures, we employ LLMs to generate contextually relevant code reasoning problems. These problems encourage the model to comprehend program structure, simulate execution paths, and perform multi-step symbolic reasoning. Moreover, for each generated problem, we create accompanying unit tests and input-output examples to provide executable validation signals, ensuring that the model not only understands code semantics during training but also demonstrates behaviorally verifiable performance. Building on this foundation, we further diversify the problems by converting them into various formats, such as output prediction, logical diagnosis, and code rewriting, and by introducing cross-language translations (e.g., Python to Java, C++ to Rust) to enhance both contextual and logical diversity. This strategy not only broadens data coverage and generalization ability but also promotes the learning of unified reasoning patterns and program semantics across different syntactic structures and type systems. As a result, it strengthens the model’s cross-language transferability and adaptability to real-world development tasks.

\subsubsection{Instruction Following Data}
\textbf{Progressive Construction of Complex Instructions}\quad
We begin by generating simple base instructions and iteratively increase their complexity by layering additional requirements related to style, format, and content. This bottom-up strategy allows for a controlled progression from basic to more intricate tasks, enabling the model to better generalize across instruction complexities.

\textbf{Result-Verifiable Instruction Generation}\quad
We construct instructions with explicitly verifiable constraints, such as length limits, required content, or structural requirements. These constraints enable automatic validation of model outputs through rule-based filtering. By generating multiple outputs under varied decoding parameters, we can efficiently identify and retain only those responses that strictly satisfy the given conditions---thus supporting scalable and accurate data generation.

\textbf{Enhancing Instruction Diversity}\quad
To enrich the diversity of the instruction set, we incorporate prompts from a wide range of domains and contexts. Inspired by the approach proposed in \citet{ge2024scaling}, we combine domain-specific knowledge with various persona settings to generate instructions that reflect a broader spectrum of real-world applications.

\textbf{Reverse Instruction Generation from Existing Data}\quad
Beyond generating instructions from scratch, we also employ a reverse instruction generation strategy---treating existing high-quality text as target outputs and generating corresponding instructions that could plausibly lead to them. This technique enables us to leverage valuable unlabeled corpora to augment instruction-following datasets.

\subsubsection{Long-Context Data}
Inspired by LongAlign~\citep{bai2024longalign}, we construct long-context supervised fine-tuning data from existing pretraining corpora. 
We sample a document $d$ from various sources in the pretraining corpus, including web pages, source code, mathematical content, encyclopedic texts, etc. For each document $d$, we use an LLM to generate a set of $n$ task-oriented queries $Q = {q_1, q_2, \ldots, q_n}$, covering a range of objectives such as extraction, summarization, reasoning, and open-domain question answering.
To simulate long-context reasoning, we retrieve related but potentially irrelevant documents to form a challenging long-context input, simulating distractor-heavy settings. For each query $q_j \in Q$, we retrieve $k$ related documents ${d_{j,1}, d_{j,2}, \ldots, d_{j,k}}$ from an indexed corpus. We then concatenate the original document $d$ with the retrieved documents to form an extended context $C_j = \text{Concat}(d_{j,1}, \ldots, d_{j,m-1}, d, d_{j,m}, \ldots, d_{j,k})$, where $d$ is inserted at a randomly selected position $m \in {1, 2, \ldots, k+1}$ within the sequence.
An LLM is prompted with each query $q_j$ alongside the context $C_j$, and tasked with generating an answer $a_j$. 

This design helps the model learn to locate relevant content and perform reasoning across long inputs with mixed relevance.
To ensure coverage across different context lengths, we control the total token count of $C_j$ to be uniformly distributed between 8K and 64K tokens.

\subsubsection{Tool Use Data}
\textbf{Function Calling}\quad
The function calling dataset combines publicly available sources such as \texttt{xlam-function-calling-60k}~\citep{liu2024apigen} and 
\texttt{glaive-function-calling-v2}~\footnote{https://huggingface.co/datasets/glaiveai/glaive-function-calling-v2}, with a substantial amount of in-house data generated via in-context learning.
To ensure data quality, we apply strict filtering criteria. Specifically, we remove samples where the tool invoked in the ground truth is not included in the available tool set, or where the parameter names or types are inconsistent with the tool schema.
Additionally, we prepend a chain-of-thought reasoning step before the tool invocation. Empirically, we find that this improves model performance by guiding it to better understand the task and select appropriate tools and arguments.

\textbf{Code Interpreter}\quad 
To enhance the model’s ability in code generation and reasoning, we construct a collection of data examples focused on solving problems with the help of a code interpreter. This includes both curated open-source datasets and internal data designed to reflect real-world coding scenarios.

For open-source datasets, we utilize resources such as CodeAct~\citep{wang2024executable} and Code-Feedback~\citep{zheng2024opencodeinterpreter}. To ensure compatibility with our execution environment, we preprocess the data by analyzing the abstract syntax tree (AST) of each code snippet and filtering out those that import external packages or rely on user interaction.

For in-house data, we collect various types of files, including CSVs, PDFs, images, and videos, and prompt an LLM to generate realistic, code-solvable problems related to the content of each file. Then the model is prompted to solve the task using a code interpreter, and is allowed to iteratively generate and execute code in a sandboxed environment. After each execution, the result is fed back to the model. If the model fails to solve the problem within 10 attempts, the data point is discarded. This approach helps create feedback-driven examples that teach the model how to use code to solve real-world tasks more effectively.

\subsection{Chunk-wise Rollout: Deep Reasoning with Load-Balanced Reinforcement Learning}
Recent research has demonstrated that RL can enhance the deep reasoning capabilities of LLMs~\citep{o1,DeepSeek-r1}. However, directly applying RL to an end-side base model often leads to unstable training and slow convergence. Thus, we first perform SFT on the base model using long-CoT distilled data. This step equips the model with basic reasoning abilities and provides a better initialization for RL. Subsequently, we proceed with RL to further enhance the model’s performance. To improve training efficiency, we carefully curate the training data and introduce a chunk-wise rollout strategy, which significantly accelerates the RL process by optimizing GPU utilization and minimizing computational waste.

\begin{table}[t]
    \centering
    \begin{tabular}{@{}p{1.0\textwidth}@{}} 
        \toprule 
        \textbf{Algorithm 1} \; Chunk-wise Rollout-based Policy Optimization \\
        \midrule 
\textbf{Input} initial policy model $\pi_\theta$; reward model $R$; task prompts $\mathcal{D}$; hyperparameters $\varepsilon_\mathtt{low}, \varepsilon_\mathtt{high}$ \\
1: Initialize replay buffer $\mathcal{R} \leftarrow \emptyset$, dynamic sampling buffer $\mathcal{B} \leftarrow \emptyset$, log-prob buffer $\mathcal{L} \leftarrow \emptyset$ \\
2: \textbf{for} step = 1,...,M \textbf{do} \\
3: \;\;\; Sample a batch $\mathcal{D}_b$ from $\mathcal{D}$ \\
4: \;\;\; \textbf{if} replay buffer $\mathcal{R}$ is not empty \textbf{then} \\
5: \;\;\;\;\;\;\; Append unfinished queries $q$ from $\mathcal{R}$ to $\mathcal{D}_b$ \\
6: \;\;\; Update old policy model $\pi_{\theta_{\text{old}}} \leftarrow \pi_\theta$ \\
7: \;\;\; \textbf{for each} $q \in \mathcal{D}_b$ \textbf{do} \\
8: \;\;\;\;\;\;\; \textbf{for} $i = 1$ to $G$ \textbf{do} \\
9: \;\;\;\;\;\;\;\;\;\; Generate a chunked output $o_i \sim \pi_{\theta_{\text{old}}}(\cdot \mid q)$ \\
10: \;\;\;\;\;\;\;\;\;\; \textbf{if} $o_i$ is unfinished: store $(q, i, o_i)$ in replay buffer $\mathcal{R}$ \\
11: \;\;\;\;\;\;\; \textbf{if} all $G$ outputs for $q$ are completed: \\
12: \;\;\;\;\;\;\;\;\;\; Compute rewards $\{r_i\}_{i=1}^{G}$ for each sampled output $o_i$ by running $R$  \\
13: \;\;\;\;\;\;\;\;\;\; Filter out $o_i$ and add the remaining to the dynamic sampling buffer $\mathcal{B}$ \Cref{eq:sample_filter}\\
14: \;\;\; \textbf{if} $|\mathcal{B}| < N$: \textbf{continue} \\
15: \;\;\; Sample a train batch $\mathcal{B}_\text{train} \subset \mathcal{B}$ of size $N$ for training \\
16: \;\;\; Let $\mathcal{B}_\text{rest} = \mathcal{B} \setminus \mathcal{B}_\text{train}$ \\
17: \;\;\; Combine $\mathcal{B}_\text{rest}$ and $\mathcal{R}$ for current-policy log-prob estimation \\
18: \;\;\; Compute $\log \pi_\theta(o_{i,t})$ for all cached chunks in $\mathcal{B}_\text{rest} \cup \mathcal{R}$ \\
19: \;\;\; Store log-probs in log-prob buffer $\mathcal{L} \leftarrow \mathcal{L} \cup \{\log \pi_\theta(o_{i,t})\}$ \\
20: \;\;\; Compute token-level advantage $\hat{A}_{i,t}$ for each sample in $\mathcal{B}_\text{train}$ \\
21: \;\;\; \textbf{for} iteration = 1, ..., $\mu$ \textbf{do} \\
22: \;\;\;\;\;\;\; Update policy $\pi_\theta$ by maximizing the objective (\Cref{eq:loss}) \\
\textbf{Output} $\pi_\theta$ \\
\bottomrule
    \end{tabular}
    \captionsetup{labelformat=empty}
    \caption{}
    \label{algo:chunk-roll}
\vspace{-1.5em}
\end{table}

\subsubsection{RL Data Curation}
We collect a large amount of high-quality data in mathematics and programming to enhance the model's reasoning ability. We found that the quality and difficulty of the data play an important role in improving the model's reasoning capabilities.

\textbf{Mathematics}\quad
We collect verifiable mathematical data primarily from DAPO, Deepscaler, Numina, Prime, and other sources. During the evaluation phase, outputs that do not conform to the expected reasoning format are assigned a reward of zero. To determine correctness, we employ a combination of rule-based matching and symbolic verification using SymPy.

\textbf{Code}\quad
The code-related data is mainly sourced from LeetCode, TACO, Kodcode, Codeforces, and similar platforms. We execute Python code within a Firejail sandbox environment\footnote{https://github.com/netblue30/firejail}. For problems with multiple test cases, the reward is calculated based on the proportion of test cases passed, with a full reward of 1.0 assigned when all test cases pass.

\textbf{Data Filtering}\quad
After data collection, we perform deduplication on both the reinforcement learning (RL) training data and the supervised fine-tuning (SFT) data using Semhash~\citep{minishlab2025semhash}. To retain more challenging samples, we use DeepSeek-R1-Distill-Qwen-1.5B to generate four predictions for each training example and filter out those for which all four predictions are correct. Since the amount of code data is significantly smaller than the math data, we upsample high-quality code samples multiple times to increase their proportion in the training set.

\subsubsection{Training Recipe}

Building upon recent advancements in the research community, we adopt a modified version of Group Relative Policy Optimization (GRPO) \citep{deepseekmath}. In addition to the original GRPO algorithm, we incorporate the following improvements:

\textbf{Dynamic Sampling}\quad
During the RL rollout phase, we filter out prompts whose responses are all right or wrong, ensuring that all prompts in the batch contribute effective gradients while maintaining a consistent batch size. This strategy mitigates high variance in the gradient, thereby improving training efficiency and stability.

\textbf{Clip-Higher Strategy}\quad
We raise the upper clipping threshold for the importance sampling ratio. This adjustment alleviates entropy collapse in later training stages and encourages greater exploration, helping the model realize its full potential.

\textbf{Token-level Policy Gradient Loss}\quad
Instead of averaging loss at the sample level, we compute the loss at the token level. This gives longer sequences a proportionally greater weight in the gradient update, which promotes the model to learn complex reasoning patterns and suppresses undesirable behaviors such as verbosity and repetition, ultimately improving training stability and generation quality.

\textbf{Overlong Sample Filtering}\quad
We exclude responses that are truncated due to length constraints from the loss computation. This prevents penalizing valid reasoning trajectories that are prematurely cut off, thereby encouraging the model to engage in deeper reasoning.


\subsubsection{Stabilized Chunk-wise Rollout}
To mitigate inference throughput degradation caused by lengthy trajectories during the rollout phase, we propose a chunk-wise rollout strategy to maximize computational resource utilization. The workflow of this strategy consists of three steps: (1) The policy model generates trajectories of a fixed chunk length for all input samples. (2) Trajectories that are either fully completed or have reached the maximum generation length are used for training. For the incomplete ones, their log probabilities are computed and stored for later use in importance sampling. (3)The unfinished trajectories are merged with the next batch of new inputs, then the process returns to step (1). By adopting this strategy, we significantly improve GPU utilization and effectively reduce computational waste caused by excessively long outputs within a single rollout iteration. The full algorithm can be found in Algorithm~\ref{algo:chunk-roll}

Since the chunk-wise rollout strategy breaks down long responses into smaller chunks across iterations, this may lead to distributional shifts in partially sampled trajectories after the policy model is updated, which can compromise training stability. To address this challenge and ensure a more stable training process, we introduce the following techniques:

\textbf{Chunk-level importance sampling}\quad
As trajectories generated through the chunk-wise rollout strategy span multiple policy model versions, we apply importance sampling at the chunk level to account for distributional differences. Each chunk is weighted independently based on its originating policy. We found this technique to be critical for maintaining stable and efficient policy optimization.

\textbf{Dual-clip}\quad
The chunk-wise strategy introduces partial off-policy rollouts, which often lead to spikes in training loss due to high variance in sampled trajectories. To mitigate this, we incorporate dual-clip \citep{ye2020mastering}, which constrains the policy update range from both directions, effectively reducing instability caused by large discrepancies in trajectory distributions.

\textbf{KL regularization with dynamic reference updates}\quad
In contrast to recent works that remove KL loss \citep{yu2025dapo,xia2025mimo}, we observe that retaining the KL penalty is essential for the stable training of chunk-wise rollouts. To avoid overly restricting the policy model's potential, we periodically update the reference model, striking a balance between training stability and model performance.

\textbf{Garble filter}\quad
Since the chunk-wise rollout strategy reuses incomplete trajectories from previous policy models, the risk of generating corrupted or incoherent text (e.g., garbled output, excessive repetition) increases. To prevent these abnormal trajectories from destabilizing training, we introduce a garble filter that detects and excludes such samples from loss computation.

\begin{table}[t]
\centering
\small
\caption{Performance Comparison between Vanilla Rollout and Chunk-wise Rollout Strategy}
\label{tab:rollout_strategies}
\begin{tabular}{lcc|cc}
\toprule
\multirow{2}{*}{\textbf{Strategy}} 
& \multicolumn{2}{c|}{\textbf{Timings}} 
& \multicolumn{2}{c}{\textbf{Performance}} \\
\cmidrule(r){2-3} \cmidrule(l){4-5}
& Step & Sampling & AIME 2024 & AIME 2025 \\
\midrule
Distill-Qwen-1.5B  & / & / & 29.79 & 23.96  \\
Vanilla              & 488.57 & 392.61 & 32.91 & 25.21 \\
Chunk-4k           &281.27  &148.14  & 32.71 &26.04  \\
Chunk-8k           & 286.31 & 173.97 & 34.79 & 26.67  \\
Chunk-16k           & 360.79 & 250.88 & 32.50 & 25.63  \\
\bottomrule
\end{tabular}
\end{table}

Based on the aforementioned techniques, for each specific question-answer pair \((q, a)\), the behavior policy \(\pi_{\theta_{\text{old}}}\) samples a group of \(G\) individual responses \(\{o_i\}_{i=1}^G\), we reformulate the original policy optimization objective to support our proposed chunk-wise rollout. The modified objective is formally defined as follows:
\begin{equation}
\mathcal{J}(\theta) = \mathcal{J}_{\text{clip}}(\theta) - \beta \cdot \mathcal{J}_{\text{KL}}(\theta),
\label{eq:loss}
\end{equation}
\begin{equation}
\begin{aligned}
\mathcal{J}&_{\text{clip}}(\theta) = \; 
\mathbb{E}_{(q, a) \sim \mathcal{D},\; \{o_i\}_{i=1}^G \sim \pi_{\theta_{c(t)}}(\cdot \mid q)} \bigg[ \;
\frac{1}{\sum_{i=1}^{G} |o_i|} \sum_{i=1}^{G} \sum_{t=1}^{|o_i|} \\ &
\begin{cases}
\min \left(
r_{i,t}(\theta) \cdot \hat{A}_{i,t},\;
\text{clip}\big(r_{i,t}(\theta),\; 1 - \varepsilon_{\text{low}},\; 1 + \varepsilon_{\text{high}}\big) \cdot \hat{A}_{i,t}
\right), & \text{if } \hat{A}_{i,t} > 0 \\[1em]
\max \left(
\min \left(
r_{i,t}(\theta) \cdot \hat{A}_{i,t},\;
\text{clip}\big(r_{i,t}(\theta),\; 1 - \varepsilon_{\text{low}},\; 1 + \varepsilon_{\text{high}}\big) \cdot \hat{A}_{i,t}
\right),\;
c \cdot \hat{A}_{i,t}
\right), & \text{if } \hat{A}_{i,t} \le 0
\end{cases}
\bigg], \\
& \quad \qquad \quad  \quad \qquad \quad    \quad \qquad \quad \text{s.t.} \quad 0< \Big|\{o_i\mid\texttt{is\_equivalent}(a,o_i)\}\Big|< G,
\end{aligned}
\label{eq:sample_filter_piecewise_final}
\end{equation}
\begin{equation}
\mathcal{J}_{\text{KL}}(\theta) = D_{\mathrm{KL}} \left( \pi_\theta \,\|\, \pi_{\mathrm{ref}} \right), \quad r_{i,t}(\theta) = 
\frac{\pi_\theta(o_{i,t} \mid q, o_{i,<t})}{\pi_{\theta_{c(t)}}(o_{i,t} \mid q, o_{i,<t})}.
\end{equation}

\subsubsection{Experimental Analysis}
We train DeepSeek-R1-Distill-Qwen-1.5B on the DAPO dataset for 150 steps to evaluate our proposed chunk-wise rollout strategy. 
The experiments are conducted on 64 A800 GPUs, using a batch size of 64 and a learning rate of $3 \times 10^{-6}$. During training, we set the number of rollouts per sample to 8. For evaluation, we report the average performance over 16 independent runs. The results are presented in Table~\ref{tab:rollout_strategies}. "Naive" refers to generating a full response for each query during the rollout phase; "Chunk-nk" denotes our chunk-wise rollout strategy, which generates up to nk tokens per input during each rollout. The metrics "step" and "sampling" represent the average time per training step and the average time spent sampling trajectories per step, respectively. All timing values are normalized with respect to the naive dynamic sampling baseline.

From the experimental results, we can observe that the chunk-wise rollout strategy effectively reduces both training and sampling time per step while maintaining performance. As the chunk size decreases, the sampling time per step steadily declines, confirming that this strategy mitigates long trajectory-induced bubbles during sampling and improves GPU utilization. However, when the chunk size is reduced from 8k to 4k, although the sampling time per step is obviously reduced, the total training time per step remains largely unchanged. This is because the smaller chunk size, while alleviating sampling bottlenecks, introduces more frequent log probability computations for chunk-level importance sampling. As a result, the overall training efficiency sees little improvement. In future work, we will further optimize this trade-off to achieve a better balance between sampling speed and log probability calculation.


\subsubsection{Implement Details}

We set the training batch size to 256 and the mini-batch size to 128. A constant learning rate of $1e-5$ is used throughout training. To accommodate the architectural design of \modelname{}, we adopt the $\mu$P learning rate strategy. Unlike recent approaches that remove the KL penalty and introduce an entropy constraint in the policy loss, we retain the KL penalty with a coefficient of 0.001 and remove the entropy constraint to ensure stable training. The maximum response length is set to 32,768 tokens, which enables the model to perform extended chain-of-thought reasoning. During the rollout phase, both the temperature and top-p are set to 1.0 to encourage broader exploration. For each query, we generate 16 rollouts to promote diversity and robustness.

\subsection{BitCPM4: Quantization-Aware Training for Ternary LLMs} \label{sectionQAT} 

Deploying LLMs is challenging due to their high computational and memory demands. Model quantization addresses this challenge by lowering parameter precision, leading to efficient inference with reduced resource consumption. Extremely low-bit quantization (e.g., 1-bit, 2-bit) has recently garnered significant interest and shows great promise~\citep{wang2023bitnet,ma2024era,xuonebit}. However, to build these extremely low-bit LLMs, PTQ methods~\citep{gptq} may not be sufficient to maintain the model performance, necessitating more effective QAT methods~\citep{liu-etal-2024-llm}. Some recent efforts, such as BitNet~\citep{ma2025bitnetb1582b4ttechnical}, even train extremely low-bit LLMs from scratch. This paper introduces an efficient QAT method to construct an effective ternary model BitCPM4, and demonstrates the feasibility of adapting high-precision LLMs to extremely low-bit versions. This means that we can greatly reduce the additional overhead caused by quantization and dequantization during QAT.

\subsubsection{Efficient Quantization-Aware Training}

For prevailing quantization methods, both weights and activations are typically quantized. Our preliminary experiments indicate that for extremely low-bit quantized LLMs, quantizing activations increases the QAT overhead without reducing too much inference costs on end-side devices. Consequently, we apply ternary quantization to model weights rather than activations. The current state-of-the-art ternary LLM, BitNet-2B~\citep{ma2025bitnetb1582b4ttechnical}, is trained from scratch using QAT on 4T tokens and achieves impressive performance. In contrast, our strategy involves initializing the ternary model with a pre-trained high-precision checkpoint to reduce the required training tokens for QAT.

To validate the feasibility of applying QAT to a model initialized from a high-precision checkpoint, we first conduct preliminary experiments on a 5M-parameter model in our ModelTunnel, using a total of 2B (400N) tokens. For these experiments, we employ the whole WSD learning rate scheduler, where the combined warm-up and stable phases account for 80\% of the total tokens, and the decay phase accounts for the remaining 20\%. These preliminary experiments involve two training stages: initially training an FP8 model, and then converting this FP8 model to a ternary model through QAT. Sufficient experiments in our ModelTunnel show that the re-warmup of the learning rate at the beginning of the second stage is critical for maintaining the performance of the FP8 model. To this end, we employ a learning rate of 1e-2 in the first stage, while in the second stage, we use a learning rate of 5e-3. While keeping the total number of training tokens constant, we adjust the allocation ratio between the FP8 training stage and the QAT stage, and record the final converged loss for each configuration. 

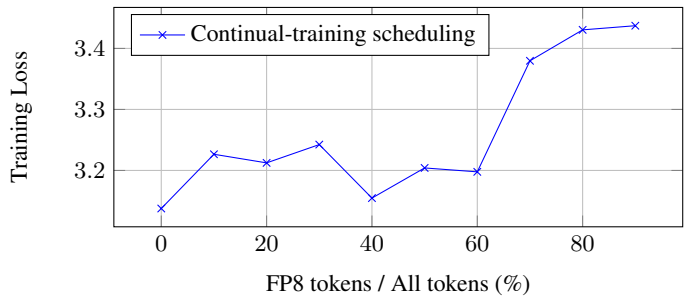
\begin{figure}[t]
    \centering
    \small
    \begin{tikzpicture}
        \begin{axis}[
            xlabel={FP8 tokens / All tokens (\%)},
            ylabel={Training Loss},
            legend pos=north west,
            grid=major,
            width=0.6\textwidth,
            height=4.5 cm
        ]
            \addplot[color=blue, mark=x] coordinates {
                (0,3.1376) (10,3.2265) (20,3.2124) (30,3.2424) (40,3.1547) (50,3.2041) (60,3.1977) (70,3.3797) (80,3.4304) (90,3.4372)
            };
            \legend{Continual-training scheduling}
        \end{axis}
    \end{tikzpicture}
    \caption{The relationship between language modeling loss and ratio of QAT post-training tokens (proportion of full stable-phase tokens).}
    \label{fig:qat_ratio}
\end{figure}

As shown in Figure~\ref{fig:qat_ratio}, when the proportion of tokens dedicated to QAT exceeds 40\% of the total training tokens, i.e., equivalent to twice the number of tokens used in the learning rate decay phase, the final loss closely approaches that of training the ternary model from scratch via QAT. Furthermore, we conduct verification experiments on a 150M-parameter model, and the results confirm that reasonable continual-training scheduling can achieve the same effect as performing QAT from scratch to get effective ternary models. To this end, we build BitCPM4 using only twice the number of tokens used in the learning rate decay phase.

\subsubsection{Discussion for Extremely Low-Bit LLMs}

We finally train two sizes of ternary models: one is BitCPM4-0.5B trained based on \modelname{}-0.5B, and the other 1B-parameter model is trained based on the internal experimental model. The whole QAT process uses 350B tokens.

We compare our models with those of other related models, and the results are shown in Table~\ref{Table:QATbenchmark}. At the 0.5B-parameter level, BitCPM4-0.5B outperforms qwen3-0.6B on knowledge-related tasks (MMLU, CMMLU, C-EVAL, etc). At the 1B-parameter level, BitCPM4-1B performs similarly to competing 2B-parameter models. As the number of tokens required for BitCPM4 is only 10\% of that for BitNet-2B, this means that implementing QAT from scratch is not necessary, and further shows that our proposed QAT method can deliver competitive results while requiring less training costs.

However, our 0.5B-parameter model exhibits relatively weaker performance on more challenging mathematical and coding tasks, which we attribute to that smaller model size restricts reasoning capabilities. Existing quantization efforts suggest that the quantization effectiveness follows a scaling law related to the model size~\citep{ouyang2024lowbitquantizationfavorsundertrained, kumar2024scalinglawsprecision}. Based on this law, we plan to apply our QAT method to larger models in our future work. Moreover, the operators for extremely low-bit models are also an issue that needs to be fully considered, and this will also be our future work.

\begin{table}[t]
\centering
\small
\caption{The comparison of BitCPM4 with other representative models. The data marked with asterisks are from the original paper of the model, and the rest of the data are reproduced by ourselves.}
\renewcommand{\arraystretch}{1.2}
\begin{tabular}{l|ccccccc}
\toprule
Model          & Qwen3 & Llama3.2 & Gemma3 & BitNet  & BitCPM4  & BitCPM4  \\ \midrule
\# Parameter   & 0.6B  & 1B       & 1B     & 2B      & 0.5B    & 1B      \\
Precision      & BF16  & BF16     & BF16   & Ternary & Ternary & Ternary \\ \midrule
MMLU           & 42.95 & 46.89    & 41.64  & 53.17*  & 49.88   & \textbf{59.24} \\
CMMLU          & 42.05 & 23.73    & 25.09  & 27.61   & 55.88   & \textbf{68.84} \\
CEval          & 45.53 & 36.74    & 31.83  & 29.36   & 57.51   & \textbf{69.06} \\
BBH            & 28.32 & 25.42    & 33.21  & 49.83   & 43.13   & \textbf{57.64} \\
GSM8K          & \textbf{61.71} & 39.76    & 61.26  & 58.63*  & 25.55   & 60.80           \\
MATH500        & \textbf{50.20} & 17.20    & 43.20  & 42.40    & 10.20   & 34.00          \\
MBPP           & 47.86 & 47.47    & 59.92  & 47.08       & 46.69   & \textbf{61.48}          \\
HumanEval      & \textbf{40.85} & \textbf{40.85}    & 42.07  & 38.40   & 29.88   & 37.20          \\ \midrule
Average        & 44.93 & 34.76    & 42.28  & 43.31   & 39.84   & \textbf{56.03}        \\ \bottomrule
\end{tabular}
\label{Table:QATbenchmark}
\vspace{-1em}
\end{table}

\section{Efficient Inference and Deployment} 
Due to strict constraints on compute, storage capacity, and power consumption of end-side devices such as mobile devices and personal computers, how to achieve efficient inference of LLMs under limited hardware resources has become a key technical challenge. In this section, we will introduce our inference system, CPM.cu, and deployment system, ArkInfer. 

\subsection{CPM.cu: Lightweight and Efficient CUDA Inference Framework}

We first develop a lightweight inference framework optimized for end-side NVIDIA chips. Beyond the basic features, such as static memory management and kernel fusion, we implement a highly efficient speculative sampling and integrate efficient sparse attention kernels for InfLLM v2.
Speculative sampling is a critical technique for accelerating LLM inference~\citep{leviathan2023fast,chen2023accelerating,medusa,li2024eagle}, particularly in resource-constrained end-side devices. This approach employs a draft-then-verify paradigm where a lightweight draft model generates candidate token sequences, which are subsequently verified by the target LLM in parallel. 
The recent advances in speculative sampling, such as EAGLE-2~\citep{li2024eagle}, utilize a tree-style drafting process.
By designing efficient attention kernels tailored for tree-based speculative sampling and implementing fused verification kernels, we achieve an optimized speculative sampling speed. We also implement an efficient InfLLM v2 sparse attention kernel within the framework.

Based on the framework, we identify that the efficiency bottleneck in speculative sampling for end-side models lies in the language modeling head of the draft model. To address this, we propose FR-Spec~\citep{zhao2025fr}, which prunes the draft model's vocabulary based on the token frequency while preserving the full vocabulary of the target model to maintain its generation quality. We further explore combining speculative sampling with 4-bit quantized GPTQ~\citep{gptq} models. To maintain model performance, we first explore an improved quantization scheme, P-GPTQ, and subsequently validate the feasibility of integrating speculative sampling with quantization in SpecMQuant~\citep{zhang2025specmqaunt} and with long-context processing.

\subsubsection{Frequency-Ranked Vocabulary Construction and Draft Verification}

The effectiveness of speculative sampling relies heavily on the efficiency of both the drafting and verification phases. Recent advances such as EAGLE-2~\citep{li2024eagle} have made remarkable progress in reducing drafting overhead through extremely lightweight architectures, by employing single-layer Transformers for the drafting process. Contemporary models tend to adopt larger vocabularies to improve tokenization efficiency, introducing significant computational overhead in the language modeling head, making the drafting process slow, although it has only a single layer.
Our investigation reveals that the transition from small to large vocabulary models substantially increases drafting time, creating a new performance bottleneck that limits the effectiveness of speculative sampling techniques.

To address this challenge, we introduce FR-Spec~\citep{zhao2025fr}, a frequency-ranked speculative sampling framework that optimizes draft candidate selection through strategic vocabulary space compression. Our approach leverages the well-documented long-tail distribution of token frequencies in natural language, where a small subset of high-frequency tokens accounts for the majority of occurrences. By restricting the draft search to a frequency-prioritized subset of tokens, FR-Spec reduces the computational overhead of the language modeling head by up to $75\%$ while maintaining the mathematical equivalence of the verification process and preserving the correctness of the final output distribution.

\begin{figure*}[t]
    \centering
    \includegraphics[width=\textwidth]{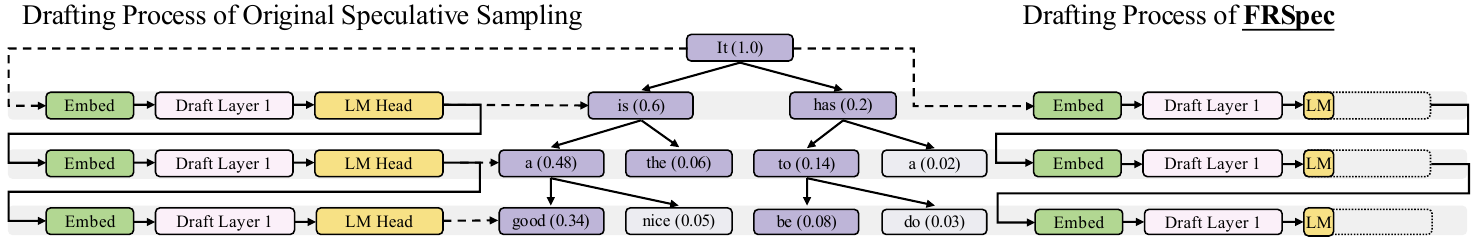}
      \caption{The illustration of FR-Spec, which requires the draft model to use a reduced vocabulary subset.}
      \label{fig:frspec}
\end{figure*}

FR-Spec introduces a frequency-ranked approach to speculative sampling that strategically optimizes the drafting phase while preserving the mathematical equivalence of the verification process. As shown in Figure~\ref{fig:frspec}, the framework operates on the principle of vocabulary space compression, where the draft model's computational scope is restricted to a subset of high-frequency tokens, thereby reducing overhead while maintaining acceptable drafting quality.

\textbf{Frequency-Based Vocabulary Subset Construction.} The foundation of FR-Spec lies in the systematic identification and selection of high-frequency tokens. We perform corpus-level analysis on large-scale pre-training data to establish comprehensive token frequency rankings. Let $f(t)$ denote the frequency of token $t$ in the corpus $\mathcal{C}$. We sort all tokens $t \in \mathcal{V}$ in descending order of frequency and select the top-$k$ tokens to form our reduced vocabulary subset:
$\mathcal{V}_{\text{high}} = \{t_1, t_2, \ldots, t_k\} \quad \text{ where } f(t_1) \geq f(t_2) \geq \cdots \geq f(t_k)$.
Our empirical analysis indicates that selecting approximately 25\% of the vocabulary ($k = 0.25 \times |\mathcal{V}|$) provides optimal performance, capturing 95\% of token occurrences while achieving substantial computational reduction.

\textbf{Modified Drafting Computation.} In standard speculative sampling, the draft model computes probability distributions over the entire vocabulary $\mathcal{V}$. FR-Spec modifies this process by restricting computations to the reduced vocabulary subset $\mathcal{V}_{\text{high}}$. Given the original language modeling head matrix $\mathbf{W}_{\text{LM}} \in \mathbb{R}^{|\mathcal{V}| \times d}$, we construct a reduced matrix $\mathbf{\tilde{W}}_{\text{LM}} \in \mathbb{R}^{|\mathcal{V}_{\text{high}}| \times d}$ by extracting rows corresponding to high-frequency tokens:
\begin{equation}
    \mathbf{\tilde{W}}_{\text{LM}}[i, :] = \mathbf{W}_{\text{LM}}[\mathcal{V}_{\text{high}}[i], :], \quad i = 1, \ldots, |\mathcal{V}_{\text{high}}|.
\end{equation}
The modified drafting computation becomes:
\begin{equation}
\mathcal{D}_{\text{FR}}(\mathbf{x}) = \text{Softmax}(\mathbf{H}_{\text{draft}}(\mathbf{x}) \mathbf{\tilde{W}}_{\text{LM}}^T),
\end{equation}
where $\mathbf{H}_{\text{draft}}(\mathbf{x}) \in \mathbb{R}^{n \times d}$ represents the hidden states from the draft model for the input sequence $\mathbf{x}$.

\textbf{Verification Process.}
A critical design principle of FR-Spec is maintaining the mathematical correctness and distributional equivalence of the verification process. The target LLM continues to operate over the complete vocabulary space $\mathcal{V}$, ensuring that the final output distribution remains identical to standard speculative sampling methods: $\mathcal{P}_{\text{target}}(\mathbf{x}) = \text{Softmax}(\mathbf{H}_{\text{target}}(\mathbf{x}) \mathbf{W}_{\text{LM}}^T)$.
This preservation guarantees that FR-Spec produces statistically equivalent results while achieving computational speedup. 

\textbf{Computational Complexity Analysis.}
The computational benefits of FR-Spec are substantial and well-defined. The language modeling head computation complexity reduces from $\mathrm{O}(nd|\mathcal{V}|)$ to $\mathrm{O}(nd|\mathcal{V}_{\text{high}}|)$, where $n$ is the draft sequence length and $d$ is the hidden dimension. The reduction factor is $\frac{|\mathcal{V}|}{|\mathcal{V}_{\text{high}}|}$.
Similarly, the softmax computation scales down proportionally, as the input dimension reduces from $\mathbb{R}^{n \times |\mathcal{V}|}$ to $\mathbb{R}^{n \times |\mathcal{V}_{\text{high}}|}$. For typical configurations where $|\mathcal{V}_{\text{high}}| = 0.25 \times |\mathcal{V}|$, this represents a $4\times$ reduction in computational overhead for the language modeling head of drafting models.

FR-Spec is designed as a plug-and-play enhancement that seamlessly integrates with existing speculative sampling techniques without requiring model retraining or architectural modifications. The method can be applied to various speculative sampling frameworks by simply replacing the standard drafting computation with the frequency-ranked variant.

\subsubsection{P-GPTQ: Prefix-Aware Post-Training Quantization for End-side Devices}

As we mentioned above, PTQ and QAT are both important approaches for model quantization. Compared with QAT, PTQ is lighter and easier to conduct. PTQ for on-device deployment necessitates simultaneous quantization of both weights and activations due to constrained computational resources. 

Recent studies demonstrate that most LLMs exhibit massive activations~\citep{sun2024massive} at initial token positions, substantially degrading activation quantization fidelity. To address this challenge, we follow the PrefixQuant method~\citep{chen2024prefixquant} to isolate these initial token activation outliers. Specifically, our solution stores initial key-value activations during preprocessing, thereby preventing excessive activation magnitudes throughout the forward propagation of the first few input tokens.

Although PrefixQuant to some extent addresses activation outliers during inference, we observe that initial token bias also significantly impacts the weight quantization calibration process. Building on this insight, we develop Prefix-Aware GPTQ (P-GPTQ), an extension of the GPTQ method~\citep{gptq} that eliminates initial token interference during Hessian computation. Formally, GPTQ computes the Hessian matrix $\mathbf{H}$ from the calibration data $\mathbf{X} \in \mathbb{R}^{n \times d}$ as
\begin{equation}
\label{eq:hessian}
\mathbf{H} = \mathbf{X}^\top \mathbf{X}.
\end{equation}

Our experimental analysis on \modelname{} shows that, when computing the covariance matrix $\mathbf{H}$ for down-projection layers, particularly in those deeper Transformer blocks, the beginning of sentence token and some initial tokens consistently introduce significant statistical bias. These initial positions exhibit activation magnitudes 10$\times$ larger than subsequent tokens, disproportionately dominating the covariance structure and leading to suboptimal quantization parameters. To mitigate this bias, P-GPTQ implements a position-aware calibration strategy. Through empirical analysis across multiple layers, we find that token positions starting from $s=4$ exhibit stable statistical features. We compute the Hessian matrix only considering stable token positions as
\begin{equation}
    \label{eq:prefix_hessian}
    \mathbf{\hat{H}} = \mathbf{X}_{\text{valid}}^\top \mathbf{X}_{\text{valid}},
\end{equation}
where $\mathbf{X}_{\text{valid}} = \mathbf{X}_{[s:]}$excludes the first $s$ positions. The idea of P-GPTQ maintains the compatibility with other quantization techniques, including rotation methods like Quarot~\citep{ashkboos2024quarot} and smoothing methods like AWQ~\citep{lin2024awq}, enabling seamless integration into existing quantization pipelines. 

We evaluate P-GPTQ and its extension under the setting where all linear layers are quantized to a per-group INT4 format. The calibration employs 1,024 randomly selected sequences from the training dataset. Table~\ref{tab:pgptq_results} presents comparative results across quantization methods, where S denotes the smoothing process applied via AWQ preprocessing. The results demonstrate that S-P-GPTQ achieves superior performance among quantized methods, exhibiting the smallest performance degradation compared to the FP16 baseline. 

\begin{table}[t]
\centering
\small
\caption{The evaluation results of different quantization methods.}
\label{tab:pgptq_results}
{\renewcommand{\arraystretch}{1.2}
\begin{tabular}{l|c|cccc}
\toprule
\textbf{Benchmark} & FP16 & GPTQ & P-GPTQ  & S-GPTQ & S-P-GPTQ  \\
\midrule
MMLU       & 75.55 & 75.01 & 75.21 & 75.05 & \textbf{75.40} \\
CMMLU      & 82.12 & 81.36 & 81.36 & 81.39 & \textbf{81.95} \\
CEval      & 81.42 & 80.08 & 80.92 & \textbf{80.71} & 80.70 \\
BBH        & 70.70 & 69.78 & 69.97 & \textbf{70.17} & \textbf{70.17} \\
GSM8K      & 82.41 & 80.71 & 80.53 & 79.83 & \textbf{80.67} \\
Math500    & 60.20 & 59.72 & 59.62 & 59.80 & \textbf{60.00} \\
MBPP       & 76.65 & 73.54 & \textbf{75.68} & 75.49 & 75.49 \\
\midrule
Average & 75.58 & 74.31 & 74.76 & 74.63 & \textbf{74.91} \\
\bottomrule
\end{tabular}}
\end{table}

\subsubsection{Speculative Sampling Meets Quantization and Long-Context}

\textbf{Quantization for Target Model}\quad
In SpecMQuant~\citep{zhang2025specmqaunt}, we provide a systematic analysis of key factors to consider when applying speculative sampling to quantized models. For example, when using EAGLE-2 with W4A16 target models (such as the model quantized by GPTQ~\citep{gptq}), the drafting process should use fewer draft tokens compared to non-quantized target models since the quantization model has alleviated the memory access bottleneck of target models. Specifically, denote the verification time of $n$ draft tokens as $T_v(n)$ and the vanilla decoding time of the target model as $T_t$, the verification-to-decoding time ratio $T_v(n)/T_t$ grows significantly faster with increasing $n$ in the quantized model compared to that in the non-quantized model. This results in the increased acceptance length from using more draft tokens being offset by the significant extension in verification time.

\textbf{Quantization for Draft Model}\quad
We further apply quantization to the EAGLE-2 draft model. This could make the drafting process even faster. Additionally, compressing the draft model to a 4-bit version also makes it suitable for memory-constrained end-side deployment. However, QSpec~\citep{zhao2024qspec} finds that using GPTQ on EAGLE-2 would lead to substantial degradation of the acceptance rate. Therefore, we change to using QAT (Section~\ref{sectionQAT}) on EAGLE-2.
We verify that the EAGLE-2 quantized by our QAT method does no harm to the average acceptance length of the speculative sampling process. 

\textbf{InfLLM v2 for Target Model}\quad
We implement the InfLLM v2 sparse attention kernel for long-context scenarios. To support tree-style draft verification in speculative sampling, assuming the total number of drafted tokens is $n$, we only construct a local 2d attention mask for the last $n\times n$ region and compress the mask via uint64 bit-packing before passing it to the InfLLM v2 kernel.

\textbf{Sliding Window for Draft Model}\quad
For long-context scenarios, if the draft model in speculative sampling uses full attention, it will significantly increase the model's first-token latency. Following the approach in TriForce~\citep{suntriforce}, we apply sliding window attention to the draft model.
Our experiments show that this not only minimizes the impact on first-token latency but also improves drafting accuracy.

\subsection{ArkInfer: Cross-Platform Deployment System}

Beyond the challenges of limited computational resources, the fragmentation of end-side chips presents another significant hurdle. This fragmentation necessitates adapting models to multiple platforms and chip types for each new model release, leading to complex adaptation and deployment. This results in a great amount of engineering effort, making it nearly impossible for models to run efficiently across all platforms.

The core of this problem lies in decoupling and efficient code reuse: How can a single technical development and engineering effort be automatically applied across multiple platforms?

To address these pain points, we propose ArkInfer, a novel cross-platform deployment system. ArkInfer is designed to overcome the fragmentation of end-side chips by providing highly efficient inference speed and serving as a versatile cross-platform compatibility layer for various model applications. To achieve this, we introduce three key solutions: (1) a cross-platform compatible architecture design, (2) reusable and efficient speculative and constrained decoding schemes, and (3) an extensible model zoo frontend.

\subsubsection{Cross-Platform Compatible Architecture Design}

ArkInfer's architectural design is fundamentally driven by the need for unified, efficient deployment across a fragmented landscape of end-side hardware. Supporting diverse platforms such as MediaTek, Nvidia, Qualcomm, and Rockchip, each with its native inference frameworks (e.g., NeuroPilot, Genie, RK-LLM, TensorRT-LLM, and llama.cpp for CPU), ArkInfer seamlessly integrates these as adaptable backends.

At its core, ArkInfer implements a powerful abstraction layer. This layer features a system of adapters that normalize the varied APIs of different backends, presenting a consistent interface to higher-level components. This ensures seamless interaction regardless of the underlying hardware or framework. Data handling is further streamlined through a unified Tensor structure, which wraps diverse data types and dimensions for consistent manipulation across the system. Critical for LLM efficiency, a dedicated KV cache manager intelligently orchestrates historical state storage and retrieval, optimizing subsequent token generation.

The central component of this architecture is an abstract executor interface, which governs the runtime execution of all model-related processes, with inputs and outputs defined by fundamental tensor types. This encompasses core neural network execution (handling the encoding of various input modalities like text, images, and audio, and autoregressive decoding), sophisticated sampling techniques for generating diverse outputs, and comprehensive preprocessing capabilities to prepare data for model input. Beyond these, ArkInfer also orchestrates complex pre-trained models by composing these fundamental executors and manages interactions with external tool-calling functionalities.

This design enables heterogeneous scheduling at the executor granularity, allowing us to fully leverage diverse computing resources. Furthermore, by tracing executor execution, we can track the flow of data and operations, which greatly facilitates debugging and performance analysis, particularly for crucial per-stage precision alignment—a common pain point in end-side adaptation.

\subsubsection{Reusable and Efficient Speculative and Constrained Decoding Schemes}

Efficient LLM inference techniques generally fall into three categories: quantization, sparsity, and acceleration of the autoregressive process. While the first two, such as GPTQ, MoE, and our InfLLMv2, are often deeply coupled with specific hardware or operator implementations, acceleration techniques like speculative sampling and constrained decoding are relatively loosely coupled with the underlying hardware. This decoupling allows us to implement these optimizations once within a deployment framework and enable them across multiple chip architectures. Therefore, ArkInfer integrates both speculative sampling and constrained decoding functionalities. Our design philosophy centers on achieving universal applicability and ease of integration within the existing execution backends. 

Central to ArkInfer's ability to generate output tokens is a core component that takes processed input and orchestrates the autoregressive generation process. This component supports a range of advanced decoding strategies to meet diverse inference needs: 

\textbf{Accelerated Speculative Decoding}\quad
For enhanced inference speed, besides the above-mentioned speculative sampling methods, ArkInfer incorporates an advanced speculative decoding mechanism based on the BiTA algorithm~\citep{lin2024bitabidirectionaltuninglossless}. This technique is a strategic choice because it dramatically boosts performance without requiring additional draft models or specialized architectural changes, simplifying deployment on resource-constrained end-side devices while maintaining high output quality.

\textbf{Constrained Decoding}\quad
To ensure outputs adhere to specific formats, such as JSON or SQL, ArkInfer employs a powerful constrained decoding method leveraging the Guidance algorithm. This approach is selected for its superior ability to enforce structural adherence and provide deterministic responses, which is crucial for applications that demand structured or precise outputs.

\subsubsection{Extensible Model Zoo Frontend}

A key hurdle in deploying models on edge-side devices stems from the fragmented nature of model file structures. Various chip manufacturers frequently impose their own distinct requirements and formats, resulting in a convoluted and inefficient deployment workflow. We contend that the optimal approach involves maintaining a centralized model zoo offering a broad selection of pre-adapted models.

To tackle this, we engineer an extensible, cross-platform frontend for ArkInfer. This interface allows users to directly access and execute various models in our model zoo, thereby notably streamlining the deployment of MiniCPM and other models across diverse devices. In addition to accelerating the growth and maintenance of our model zoo, we also create an automated model conversion pipeline. This system can efficiently convert models into the formats required by different platforms, greatly accelerating the ongoing development of our model zoo.

\section{Evaluations}
Based on the efficient training and inference mechanism, we bulid \modelname{}-8B and \modelname{}-0.5B. In this section, we evaluate the effectiveness and efficiency of our models on several open-source benchmarks.

\subsection{Experimental Settings}
\textbf{Benchmarks}\quad
\modelname{} was primarily pre-trained on Chinese and English corpora. Therefore, we select the following datasets to evaluate our model, including knowledge-intensive evaluation sets MMLU~\citep{hendrycksmeasuring}, CMMLU~\citep{li2024cmmlu}, and CEval~\citep{huang2023c} for English and  Chinese, and reasoning evaluation sets including general reasoning BigBench Hard~(BBH)~\citep{suzgun2023challenging}, mathematical reasoning GSM8K~\citep{cobbe2021training}, MATH500~\citep{hendrycks2measuring}, and AIME~\citep{aime}, and code reasoning MBPP~\citep{austin2021program}, HumanEval~\citep{chen2021evaluating}, and LiveCodeBench~(LCB)~\citep{livecodebench}. 
We adopt OpenCompass~\citep{2023opencompass} as our evaluation framework. 

\textbf{Baseline Models}\quad
We compare \modelname{}-8B and \modelname{}-0.5B with several widely-adopted open-source LLMs. Specifically, for \modelname{}-0.5B, we select several models with approximately 1 billion parameters, including Qwen3-0.5B~\citep{yang2025qwen3}, Llama3.2-1B~\citep{llama3.1}, Gemma3-1B~\citep{team2025gemma}. These models are well-trained with trillions of tokens and trained with knowledge distillation, which ensures their effectiveness. For \modelname{}-8B, we select models with approximately 10 billions parameters as baselines, including Qwen3-8B~\citep{yang2025qwen3}, GLM4~(0414 version)~\citep{glm2024chatglm}, Gemma3-12B~\citep{team2025gemma}, and Phi4-14B~\citep{abdin2024phi}.

\textbf{Pre-training Pipeline}\quad
We adopt $\mu$P as our basic model architecture. Before model pre-training, we first search for hyperparameters, including learning rate, batch size, and parameter initial settings, with models containing millions of parameters. Then we follow a four-stage pipeline to pre-train \modelname{}. First, we conduct a stable pre-training stage using $7$ trillion tokens with a learning rate as $7\times10^-3$.  Then we perform an annealing pre-training stage using $1$ trillion tokens. For these two stages, the context length is set as $4$K. To enable long-sequence processing, we extend the context window from $4$K to $32K$. In this stage, we train our model with $20$ billion tokens and use LongRoPE~\citep{longrope} as our position encoding. Notably, though we only train our model in $32$K context, \modelname{} can process $128$K sequences with YaRN~\citep{yarn}. 
Following three pre-training stages, we conduct hybrid supervised fine-tuning and reinforcement learning to construct a hybrid reasoning model, \modelname{}.1.
To achieve ultimate inference speedup, we further conduct additional training for the speculative head to improve the acceptance length of the draft model. 


\begin{table}[t]
    \centering
    \small
    \caption{Evaluation results of \modelname{} and other open-source LLMs.}
    \label{tab:main_results}
    {\renewcommand{\arraystretch}{1.2}
    \setlength\tabcolsep{1mm}{
    \begin{tabular}{l|cccc|ccccccc}
    \toprule
    Models & Qwen3 & Llama3.2 & Gemma3 & \modelname{} & Qwen3 & GLM4 & Gemma3 & LLaMA3.1 & Phi4 & \modelname{} \\ \midrule
    \# Parameter & 0.6B & 1B & 1B & 0.5B & 8B & 9B & 12B & 8B & 14B & 8B \\
    \# Train Data & 36T & 9T & 2T & 1T  & 36T & 10T & 12T & 15T & 10T & 8T \\ 
    \midrule
    MMLU  & 42.95 & 46.89 & 41.64 & \textbf{55.55} 
            & 77.55 & 75.90 & 73.36 & 69.38 & \textbf{81.61} & 75.83       \\
    CMMLU   & 42.05 & 23.73 & 25.09 & \textbf{65.22} 
            & 77.58 & 74.49 & 62.52 & 54.41 & 67.56 & \textbf{80.62}    \\
    CEval   & 45.53 & 36.74 & 31.83 & \textbf{66.11} 
            & 80.35 & 74.09 & 62.23 & 52.66 & 64.28 & \textbf{81.36}    \\
    BBH   & 28.32 & 25.42 & 33.21 & \textbf{49.87} 
            & 69.43 & 61.36 & 66.66 & 44.34 & 72.79 & \textbf{76.73}     \\
    GSM8K   & \textbf{61.71} & 39.76 & 61.26 & 52.08 
            & 93.25 & 89.39 & 94.16 & 84.08 & \textbf{94.77} & 91.51   \\
    MATH500 & \textbf{50.20} & 17.20 & 43.20 & 29.60 
            & \textbf{83.20} & 66.00 & 82.20 & 48.20 & 79.60 & 78.60             \\
    MBPP  & 47.86 & 47.47 & \textbf{59.92} & 59.14 
            & 77.04 & 74.71 & \textbf{84.44} & 68.09 & 80.54 & 78.99    \\
    HumanEval & 40.85 & 40.85 & 42.07 & \textbf{46.34} 
            & 85.98 & 82.32 & 83.54 & 70.73 & \textbf{86.59} & 85.37 \\ \midrule
    Average & 44.93 & 34.76 & 42.28 & \textbf{52.99} 
            & 80.55 & 74.78 & 76.14 & 61.49 & 78.47  & \textbf{81.13} \\
    \bottomrule
\end{tabular}
}}
\end{table}

\begin{table}[t]
    \centering
    \small
    \caption{Evaluation results of \modelname{}.1 and other open-source LLMs for deep reasoning tasks.}
    \label{tab:main_results_deep_reasoning}
    {\renewcommand{\arraystretch}{1}
    \setlength\tabcolsep{1mm}{
    \begin{tabular}{l|ccccc|cc}
    \toprule
    Models & Qwen3 & R1-Qwen3 & GLM-Z1 & MiMo-0530 & Nemotron-Nano-v2 & \modelname{}.1 & \modelname{}.1  \\
    \# Parameter & 8B & 8B & 9B & 7B & 9B & 8B & 8B  \\ 
    Attention    & Full & Full & Full & Full & Full & Full & Sparse \\ \midrule
    \multicolumn{8}{c}{Knowledge} \\ \midrule
    MMLU          & 86.05 & 85.36	& 85.01 & 81.54 & 84.57 & 86.38 & 86.66 \\
    MMLU-Redux    & 87.33 & 86.25	& 86.50 & 82.62 & 84.98	& 86.41	& 86.05	\\
    CMMLU         & 81.68 & 80.53	& 76.00 & 66.10 & 61.59	& 84.94	& 84.72	\\
    CEval         & 84.84 & 84.44	& 77.11 & 66.99 & 63.62	& 84.38	& 85.75	\\ \midrule
    \multicolumn{8}{c}{Math}  \\ \midrule
    GSM8K         & 95.30 & 93.86 & 95.91 & 96.13 & 95.22 & 94.16 & 94.01	\\
    MATH500       & 96.40 & 97.20 & 96.00 & 97.20 & 95.80 & 95.60 & 97.40\\
    AIME24        & 73.33 & 83.33 & 75.62 & 78.12 & 71.67 & 83.33 & 80.83	\\
    AIME25        & 66.67 & 75.21 & 55.42 & 72.50 & 56.67 & 73.33 & 72.08	\\ \midrule
    \multicolumn{8}{c}{Code}  \\ \midrule
    HumanEval     & 93.90 & 95.73 & 95.12 & 95.73 & 93.90 & 95.73 & 91.46	\\
    MBPP          & 81.32 & 92.61 & 89.49 & 91.83 & 93.39 & 92.22 & 91.05	\\
    LCB-v5        & 56.89 & 62.87 & 49.10 & 59.88 & 68.26 & 58.68 & 56.89	\\
    LCB-v6        & 48.57 & 53.14 & 42.29 & 52.00 & 60.00 & 52.00 & 51.43	\\
    MultiPL-E     & 59.22 & 54.09 & 49.48 & 53.26 & 57.50 & 57.73 & 56.84	\\ \midrule
    \multicolumn{8}{c}{Other} \\ \midrule
    IFEval        & 84.66 & 74.12 & 80.59 & 58.04 & 86.69 & 75.23 & 77.45	\\
    BBH           & 74.17 & 76.99 & 75.48 & 73.42 & 74.28 & 82.40 & 82.68\\ \midrule
    Average       & 78.02 & 79.72 & 75.27 & 75.02 & 76.54 & 80.17 & 79.69  \\
    \bottomrule
\end{tabular}
}}
\end{table}

\subsection{Standard Evaluation}
We show the evaluation results of \modelname{} and baseline models in Table~\ref{tab:main_results}. From the results, we can observe that:
1) Both of our models achieve state-of-the-art performance among models of similar size, demonstrating the effectiveness of our training strategies. Furthermore, our models outperform several open-source large language models with significantly more parameters. For instance, \modelname{}-0.5B achieves superior performance compared to Llama3.2-1B and Gemma3-1B, despite these models having twice the parameter scale of \modelname{}. Similarly, \modelname{}-8B surpasses Gemma3-12B and Phi4-14B. This further validates that by leveraging high-quality data and efficient learning algorithms, \modelname{} can achieve exceptional performance.
2) Compared to these open-source models, \modelname{} achieves excellent performance with significantly lower training costs. Specifically, \modelname{} demonstrates comparable performance to Qwen3, while Qwen3 utilizes $36$ trillion tokens for training compared to \modelname{}'s 8 trillion tokens -- representing only 22\% of Qwen3's training data scale.
3) Among the baseline models, including Qwen3-0.6B/8B, Llama3.2-1B, and Gemma3-1B/12B, employ knowledge distillation training strategies, using larger teacher models to guide the training of end-side models. Our experimental results show that despite using only ground-truth as the supervision signals, our model still exhibits strong performance. The distillation process requires substantial computational resources to deploy teacher models. In the future, we will explore more efficient model distillation strategies to further enhance the performance of our end-side models.

We present the results of \modelname{}.1-8B in Table~\ref{tab:main_results_deep_reasoning}. From the results, we can observe that:
1) \modelname{}.1-8B achieves competitive performance with an overall average score of 79.93, outperforming similar-sized models.  It demonstrates strong deep reasoning capabilities across various tasks. 
2) The comparison between full attention (79.93 average) and sparse attention (79.14 average) shows only a 0.79-point difference, indicating that sparse attention has negligible impact on model performance while providing significant computational efficiency gains. This makes the sparse variant an attractive option for practical deployment where inference speed is crucial.

\begin{figure}
    \centering
    \includegraphics[width=\linewidth]{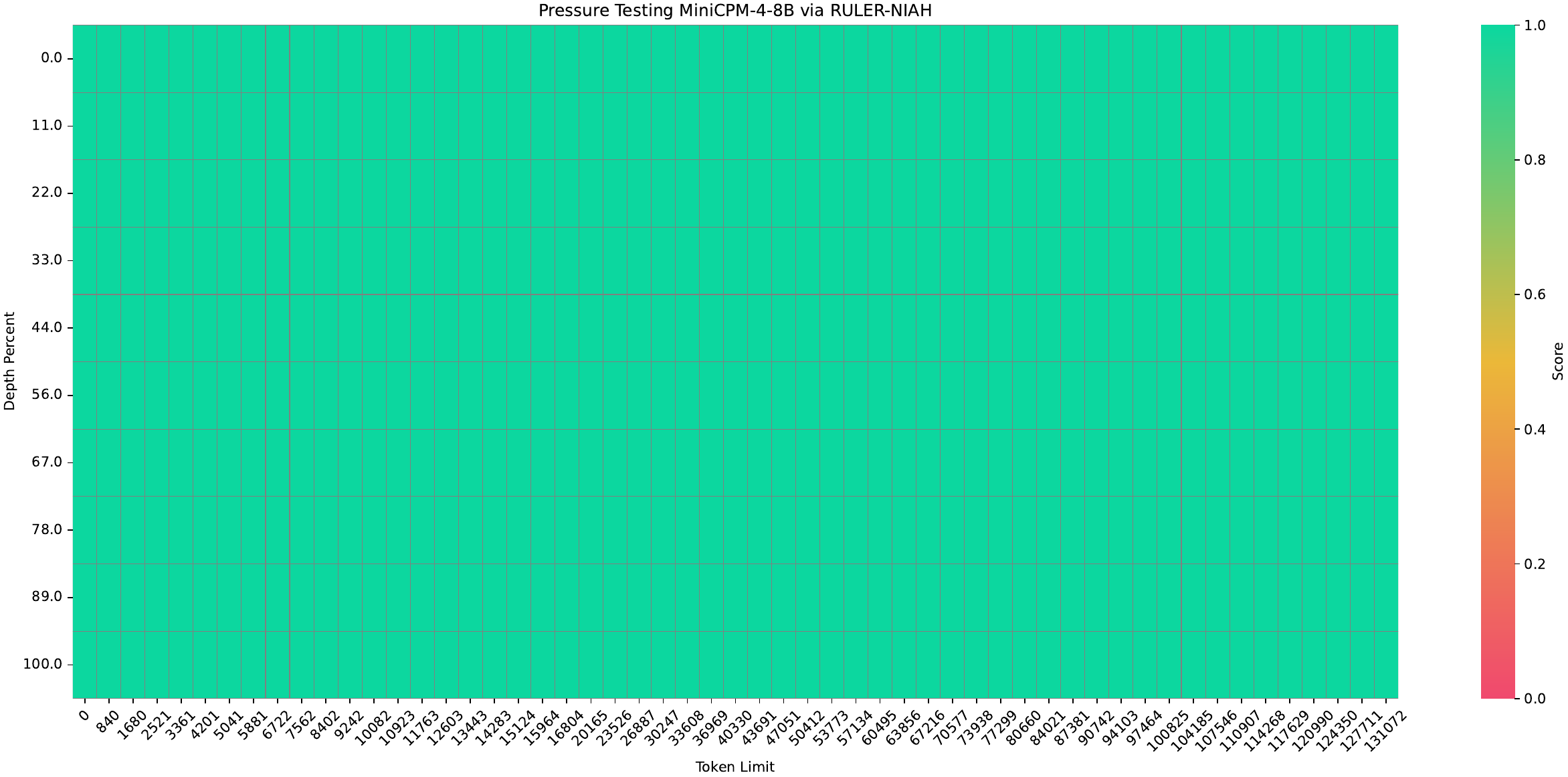}
    \caption{The evaluation results for long sequence prefilling with sparse attention.}
    \label{fig:niah}
    \vspace{-1em}
\end{figure}
\subsection{Long-Context Evaluation}
In \modelname{}, we extend the context window to $32$K using the sparse attention mechanism. In this paragraph, we evaluate \modelname{} on long sequence understanding task. Specifically, we follow \cite{hsiehruler} and evaluate our model on the needle in a haystack task (RULER-NIAH). We apply YaRN~\citep{yarn} to extend the context window of \modelname{} to $128$K and evaluate \modelname{} with $128$K NIAH.

The results are shown in Figure~\ref{fig:niah}. From the results, we can observe that: 
1) \modelname{} can achieve satisfactory performance on long sequences and achieve $100\%$ accuracy on the needle in a haystack task. And for each token, \modelname{} only requires the model to attend $6$K context tokens, which means on $128$K context, the sparsity of \modelname{} is only $5\%$.
2) \modelname{} has good performance on context window extrapolation. Even we only pre-train the model on $32$K context, \modelname{} can achieve $100$\% accuracy on $4\times$ context length.
In the following sections, we apply \modelname{} on the survey generation task, which requires the model to read and write long documents. And \modelname{} can achieve better performance than other baseline models, showing the effectiveness of \modelname{} on long-sequence processing.

\begin{table}[t]
    \centering
    \small
    \caption{Evaluation results of \modelname{}.1 on RULER~(32K).}
    \label{tab:ruler_results}
    {\renewcommand{\arraystretch}{1}
    \setlength\tabcolsep{1mm}{
    \begin{tabular}{l|ccccccccc|c}
    \toprule
    Datasets & NIAH-S & NIAH-MK & NIAH-MV & NIAH-MQ & QA1 & QA2 & VT & CWE & FWE & wAvg. \\ \midrule
    \modelname{}.1 (Full) & 100.00 & 100.00 & 91.50 & 99.50 & 76.00 & 54.00 & 85.20 & 62.60 & 87.33 & 88.93 \\
    \modelname{}.1 (Sparse) & 100.00 & 87.33 & 94.50 & 98.50 & 72.00 & 56.00 & 85.20 & 60.40 & 87.33 & 85.84 \\
    \bottomrule
\end{tabular}
}}
\end{table}

Based on the experimental results in Table~\ref{tab:ruler_results}, MiniCPM4.1 demonstrates strong long-context performance on the RULER benchmark. The sparse attention variant offers several advantages: 1) Maintained competitive performance: Despite a modest 3.09 percentage point decrease in weighted average (85.84\% vs 88.93\%), sparse attention can achieve comparable results on many long sequence tasks. 2) Computational efficiency: The minimal performance trade-off suggests that sparse attention effectively preserves long-context understanding capabilities while providing significant computational savings, making it particularly suitable for resource-constrained applications requiring extended sequence processing.

\subsection{Efficiency Evaluation}
To achieve ultimate inference acceleration, we construct a sparse attention mechanism, InfLLM v2, in \modelname{}, employ speculative sampling algorithms, FR-Spec, propose prefix-aware quantization algorithms, and build our specialized inference framework to realize ultimate speedup on end-side devices. To validate the effectiveness of our proposed algorithms, we test our model's efficiency on two typical end-side chips in this section. Specifically, we select two edge chips: Jetson AGX Orin and RTX 4090. The former is widely deployed in end-side scenarios such as automotive chips and robotics, while the latter is primarily used as computing equipment in personal computers.

The evaluation results are shown in Figure~\ref{fig:efficiency-eval}. We evaluate the throughput speed of Llama3-8B~\citep{llama3.1}, GLM4-9B~\citep{glm2024chatglm}, Qwen3-8B~\citep{yang2025qwen3}, and \modelname{} on sequences ranging from $32$K to $128$K. From the results, we can observe that:
1)~Compared with open-source LLMs with similar parameter size, we can achieve consistent speedup in both prefilling and decoding scenarios. Specifically, compared to Qwen3-8B, we achieve approximately 7x decoding acceleration on Jetson AGX Orin. The results demonstrate the effectiveness of our approach.
2)~As the text length increases, the efficiency advantage of our model becomes more pronounced. This is because the sparse attention mechanism can effectively reduce the computational and memory access overhead for long texts. As the text length that the model needs to process gradually increases, the memory access overhead of traditional dense attention mechanisms grows rapidly, while the number of context blocks that InfLLM v2 needs to access remains constant, with only the representation of semantic kernels growing slowly with sequence length. Therefore, in long sequence processing, \modelname{} can consistently handle long texts efficiently.






\section{Applications}  

The efficiency of \modelname{} unlock compelling capabilities across diverse scenarios. We highlight three key applications: 
(1) \textbf{Trustworthy Survey Generation} demonstrates its strength in efficient long-sequence processing, crucial for understanding and synthesizing complex information from large documents to produce accurate summaries.
(2) \textbf{Tool Use with Model Context Protocol} is pivotal for agent-centric deployments, enabling \modelname{} to reliably interpret instructions, context, and state information to seamlessly interact with external tools and APIs – essential for building robust, capable agents.

\subsection{\modelname{}-Survey: Trustworthy Survey Generation}

\begin{figure}[t]
    \centering
    \includegraphics[width=0.95\textwidth]{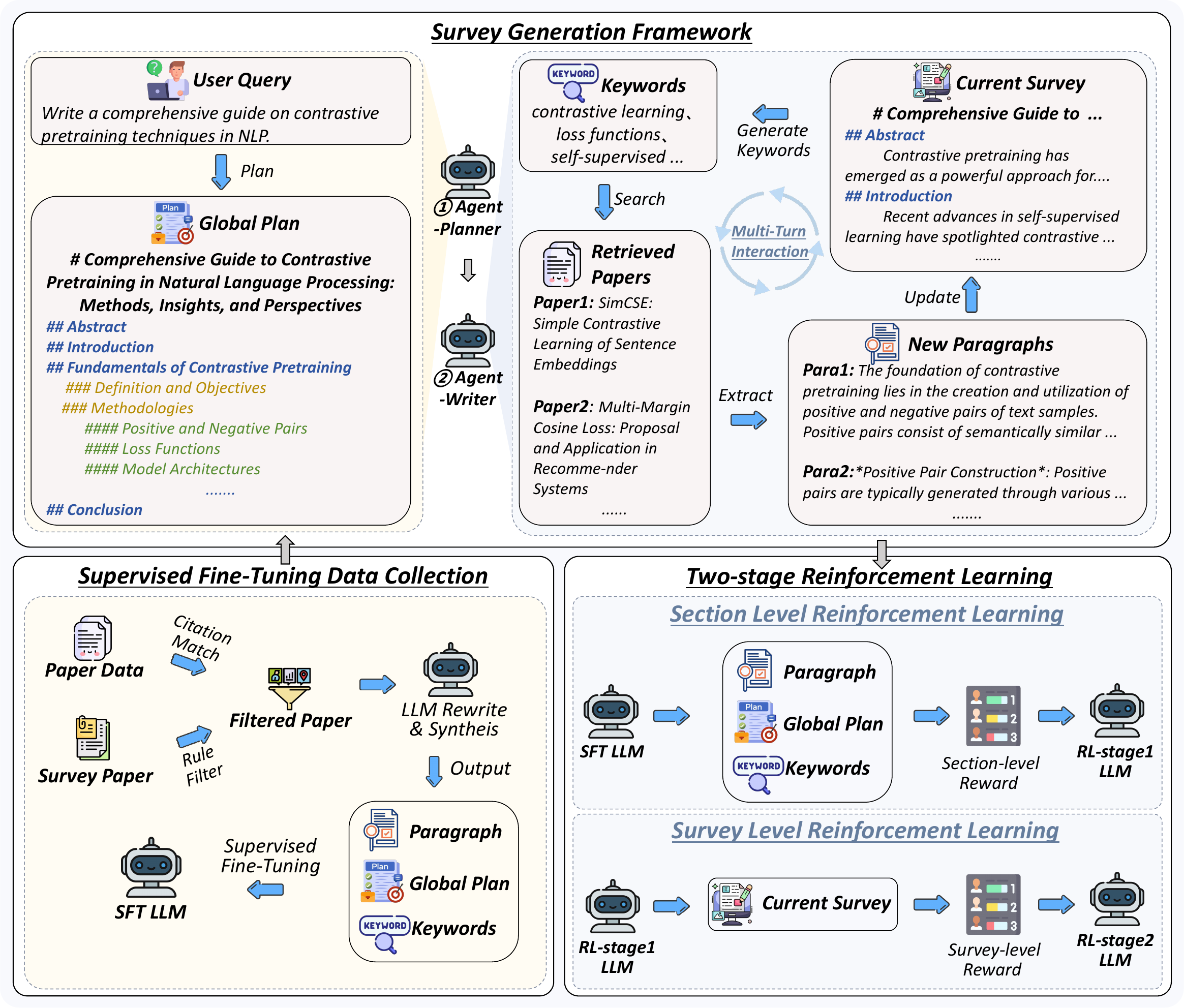} 
    \caption{The outline of~\surveyagent.}
    \label{fig:surveycpm-main} 
\end{figure}

Developing a comprehensive literature survey presents a significant challenge, even for experienced researchers. This endeavor necessitates sophisticated capabilities, including collecting pertinent resources, aggregating and synthesizing diverse information, pinpointing critical challenges, and forecasting future research trajectories. Fortunately, the swift evolution of AI technologies, spurred by LLMs, is rendering automated deep research increasingly feasible. Recent efforts~\citep{wang2024autosurvey,li2025webthinker,wang2025mapReduce}, such as projects like OpenAI Deep Research\footnote{\url{https://openai.com/index/introducing-deep-research}} and Gemini Deep Research\footnote{\url{https://gemini.google/overview/deep-research}}, leverage the long-form reasoning capabilities of modern LLMs to construct practical survey systems, simplifying the process for human researchers to assimilate extensive academic literature and rapidly grasp new fields of study.

On-device survey systems present important advantages complementary to cloud-based services. Crucially, for users with strict confidentiality needs who cannot upload resources to the cloud, locally deployed, in-house models are the only viable option. This maintains data integrity and control. Furthermore, deploying smaller-scale models on-device significantly cuts computational costs during inference, a key consideration given the high token consumption inherent in the survey writing process. This makes localized systems both secure and economically sensible for many applications.

To this end, we propose \textbf{\surveyagent}, a model built upon \modelname{}-8B that is capable of generating trustworthy, long-form survey papers while maintaining competitive performance relative to significantly larger models. Specifically, our model works in a Plan-Retrieve-Write manner, wherein it operates through three core stages: (1) {\em Planning}: defining the overall structure of the survey from a global perspective, specifying the content to be addressed in each section, subsection, and paragraph; (2) {\em Retrieval}: generating appropriate retrieval keywords based on the planner-generated outline and querying a knowledge base to obtain relevant literature; and (3) {\em Writing}: synthesizing the retrieved information to generate coherent section-level content, iteratively proceeding until the entire survey is complete.

To augment the capabilities of \modelname{}-8B for this task, we curate and process a large corpus of expert-authored survey papers to construct a high-quality training dataset. Concurrently, we compile an extensive collection of research papers to build a retrieval database. We further introduce a multi-stage training pipeline comprising: supervised fine-tuning, section-level reinforcement learning, and survey-level reinforcement learning. By combining the efficiency of a compact model with a rigorous training methodology, \surveyagent delivers performance on par with, and in some cases exceeding, that of large-scale LLMs, while remaining accessible and cost-efficient.

\subsubsection{Data Construction} 



As previously outlined, the survey generation pipeline involves several key stages, including planning, iterative retrieval, and content generation, which necessitate training data covering each of these stages comprehensively. To improve the survey generation capabilities of \modelname{}, we prioritized constructing high-quality datasets derived primarily from academic surveys, ensuring robust training outcomes. Specifically, we collected approximately 2.71 million paper abstracts from Kaggle\footnote{\url{https://www.kaggle.com/api/v1/datasets/download/Cornell-University/arxiv}} as foundational data. Subsequently, we built an efficient retrieval index using \texttt{MiniCPM-Embedding-Light}\footnote{\url{https://huggingface.co/openbmb/MiniCPM-Embedding-Light}} in conjunction with Faiss\footnote{\url{https://github.com/facebookresearch/faiss}}.

To ensure the usability and quality of the data, raw data underwent rigorous preprocessing steps before constructing the training dataset. These steps included filtering out multimodal elements, such as tables and images, removing references not indexed in the database, and standardizing data formats. It is important to note that, beyond simple rule-based methods, we significantly utilized Large Language Models (LLMs) to rewrite and refine the raw data, thereby enhancing its consistency and relevance. Additionally, adhering to the survey generation framework proposed by \cite{wang2025mapReduce}, we synthesized three distinct types of data—queries, plans, and retrieval keywords—from the preprocessed data. These synthesized datasets were systematically structured according to the Query2Plan and Plan2Survey stages, yielding 3,750 and 61,684 training samples respectively. 

\subsubsection{Training Strategy}

Given that survey generation poses considerable complexity for foundational models, we first employed small-scale Supervised Fine-Tuning (SFT) with a limited dataset to achieve effective cold-start performance and boost the likelihood of positive example sampling. Subsequently, we transitioned to a Reinforcement Learning (RL) phase, enabling steady and sustained performance enhancement. Importantly, we decomposed the RL process into two distinct phases: initially optimizing rewards at the chapter-level to ensure contextual coherence and structural precision within each chapter, followed by shifting the optimization target towards overall survey-level objectives, enhancing global coherence, depth, and thematic relevance. This progressively challenging optimization strategy significantly enhanced the final performance of the model.

It is important to highlight three primary challenges encountered during training for survey generation capabilities. First, the quality of generated surveys is difficult to reliably assess using traditional metrics such as perplexity (PPL), necessitating the creation of a dedicated reward system. Second, survey generation involves multiple complex stages, demanding robust context management to ensure retention of essential information while facilitating efficient reasoning. Third, the iterative retrieval and evaluation inherent in the generation process demand low-latency interactions to maintain efficient RL training cycles. To address these challenges, we specifically implemented the following optimizations:

\begin{table}[t]
  \small
  \centering
    \caption{Reward System for different agent abilities.}
  \setlength\tabcolsep{1.5mm}{
  \begin{tabular}{llcl}
    \toprule
    \textbf{Agent Ability} & \textbf{Metrics} & \textbf{Judgement} & \textbf{Description}\\
    \toprule
    \multirow{2}{*}{Planning} 
        & Structure Rationality & LLM \& Rule & Whether the outline is reasonable. \\
        & Structure Similarity & Rule & Whether similar to the golden plan. \\
        & Truthfulness & LLM & Whether the content in the plan is real and reliable. \\
    \midrule
    \multirow{1}{*}{Searching} 
        & Recall Score & LLM & Whether the recalled papers include the golden papers. \\
    \midrule
    \multirow{7}{*}{Content Writing} 
        & Length & Rule & Whether the length of each section is reasonable. \\
        & Language (en) & Rule & Whether the response is in English only. \\
        & Relevance & LLM & Whether focus on the user's query.\\
        & Coverage & LLM & Whether covers a wide range of related topics.\\
        & Depth & LLM & Whether reflects deep and dynamic discourse.\\
        & Novelty & LLM & Whether covers several new aspects of the user's query.\\
        & Redundancy & LLM & Whether concise and no repeat sections.\\
    \midrule
    \multirow{2}{*}{Citation Writing} 
        & Hallucination & Rule & Whether all the citations appear in the retrieved information.\\
        & Fact Score & LLM & Whether the fact claims are consistent with citations.\\
    \bottomrule
  \end{tabular}}
  \label{tab:reward_system}
\end{table}

\textbf{Reward Design:} Recognizing the crucial role reward design plays in reinforcement learning, we established a comprehensive reward framework to assess key model capabilities, including planning, retrieval, content creation, and citation accuracy. This system evaluates model performance across multiple metrics—such as structural coherence, authenticity, recall precision, relevance, content depth, coverage breadth, novelty, redundancy, hallucination frequency, and factual accuracy—leveraging both large-scale language models and rule-based evaluations. Detailed descriptions can be found in Table~\ref{tab:reward_system}.

\textbf{Context Manager:} The inherent complexity of multi-step reinforcement learning requires robust, dynamic context management. Our context manager incorporates three essential functionalities:

\begin{itemize}
    
\item Prompt Updating: Dynamically adjusts prompts according to historical interactions.

\item Historical Recording: Maintains a detailed record of interactions—including prompts, responses, penalties, retrieval actions, and evaluation scores—enabling precise reconstruction of learning trajectories.

\item Advantage Allocation: Assigns finely-tuned advantage scores through a combination of step-level format penalties and trajectory-level rewards, thus facilitating accurate token-level optimizations.
\end{itemize}

\textbf{Parallel Environment Interaction:} To mitigate latency caused by asynchronous external dependencies such as retrieval and evaluation systems, we employed parallel environment interactions using dedicated server instances. Drawing inspiration from recent developments in asynchronous API-driven multi-turn RL systems, this strategy markedly enhanced training efficiency by reducing feedback loop bottlenecks.

\subsubsection{Evaluations}




\paragraph{Baselines} 
To validate the effectiveness of our model, we also examine its performance relative to the following representative baseline systems. These systems were chosen to cover a spectrum of established approaches and provide a comprehensive benchmark for comparison: (1)\textit{Naive RAG} is a straightforward retrieval-augmented generation method. It directly inputs all query-retrieved documents to the model, which then generates a comprehensive survey in a single, non-iterative pass. (2) \textit{AutoSurvey}~\citep{wang2024autosurvey} is a structured framework for automated academic survey generation, typically employing systematic processes such as planning, multi-source literature retrieval, and coherent content synthesis for its outputs.
(3) \textit{WebThinker}~\citep{li2025webthinker} is a deep research framework powered by large reasoning models for tasks like QA or report generation. We evaluate training-free configurations using QwQ-32B~\citep{qwq32b} and the DeepSeek-R1-Distill-Qwen-7B model as comparative baselines.
(4) \textit{OpenAI Deep Research} is an OpenAI system employing multi-step online information acquisition to generate detailed reports from user queries, utilizing long-form reasoning capabilities of GPT.
Note that both AutoSurvey and Naive RAG are training-free methods. We use \texttt{gemini-2.0-flash-thinking- exp-1219}\footnote{\url{https://ai.google.dev/gemini-api/docs/models}} as the backbone for them.



\textbf{Evaluation Details}\quad
We use the SurveyEval dataset released by \citet{wang2025mapReduce} as the test set, which includes 20 test examples.
Inspired by STORM~\citep{shao2024assisting} and FactScore~\citep{min2023factscore}, we use the following four metrics to assess the quality of model-generated surveys: (1) {\em Relevance} assesses whether the survey effectively maintains clear focus on, and direct relevance to, the user's specific query, avoiding unrelated tangents or digressions. (2) {\em Coverage} evaluates whether the survey offers comprehensive coverage of the designated topic, thoroughly exploring its key sub-areas, diverse facets, and important established knowledge within the field. (3) {\em Depth} determines whether the survey thoroughly examines the core topic and its related areas, providing sufficient analytical detail, critical evaluation, and meaningful insight into complex underlying issues. (4) {\em Novelty} judges whether the survey introduces genuinely novel perspectives, original interpretations, or significant previously unarticulated connections pertinent to the user's initial query. (5) {\em Fact Score} quantifies the proportion of discrete, verifiable atomic facts presented within the survey that are accurately and explicitly substantiated by appropriate and credible cited references.
We use the GPT-4o as a judge to evaluate these metrics. 

\begin{table}[t]
  \centering
    \caption{Performance comparison of the survey generation systems. ``G2FT'' stands for Gemini-2.0-Flash-Thinking, and ``WTR1-7B'' denotes Webthinker-R1-7B. FactScore evaluation was omitted for Webthinker, as it does not include citation functionality, and for OpenAI Deep Research, which does not provide citations when exporting the results.}
  \resizebox{\textwidth}{!}{
  \begin{tabular}{lcccccc}
    \toprule
    \multirow{2}{*}{\textbf{Method}}  &  \multicolumn{5}{c}{\textbf{Content Quality}} & \multicolumn{1}{c}{\textbf{Faithfulness}} \\
     & Relevance &  Coverage & Depth & Novelty & Avg. & Fact Score\\
    \cmidrule(lr){1-1} \cmidrule(lr){2-6} \cmidrule(lr){7-7}
    Naive RAG {\small (driven by G2FT)} & 3.25  & 2.95  & 3.35  & 2.60  & 3.04  & 43.68 \\
    AutoSurvey {\small (driven by G2FT)} & 3.10  & 3.25  & 3.15  & \textbf{3.15}  & 3.16  & 46.56 \\
    Webthinker {\small (driven by WTR1-7B)} & 3.30  & 3.00  & 2.75  & 2.50  & 2.89  & -- \\
    Webthinker {\small (driven by QwQ-32B)} & 3.40  & 3.30  & 3.30  & 2.50  & 3.13  & -- \\
    OpenAI Deep Research {\small (driven by GPT-4o)} & 3.50  & \textbf{3.95}  & 3.55  & 3.00  & \textbf{3.50}  & -- \\
    \cmidrule(lr){1-1} \cmidrule(lr){2-6} \cmidrule(lr){7-7}
    \surveyagent & 3.45  & 3.70  & \textbf{3.85}  & 3.00  & \textbf{3.50}  & \textbf{68.73} \\
    \quad \textit{w/o} RL & \textbf{3.55}  & 3.35  & 3.30  & 2.25  & 3.11  & 50.24 \\
    \bottomrule
  \end{tabular}
  }

  \label{tab: results}
\end{table}

\textbf{Results}\quad
As shown in Table~\ref{tab: results}, our proposed method surpasses baseline systems driven by both open-source (e.g., Webthinker) and closed-source models (e.g., AutoSurvey) in content-related metrics. Our approach achieves performance comparable to OpenAI Deep Research, highlighting its effectiveness and competitiveness. Furthermore, our method attains the highest scores among the examined systems for factual metrics.
Furthermore, the \surveyagent shows significant improvement from SFT to RL, underscoring the efficacy of the RL component introduced in the latter stages. Specifically, enhancements are observed in Coverage, Depth, and Novelty, indicating strengthened exploration and planning capabilities. Despite these gains, \surveyagent still lags behind some baseline methods in Coverage and Novelty, suggesting further opportunities to enhance its strategic planning and exploratory capacity.
\subsection{\modelname{}-MCP: Tool Use with Model Context Protocol} 
The interaction logic between large language models (LLMs) and external tools has traditionally been statically designed and lacks standardization, which is not compatible with the rapid, independent evolution of agents and tools. This incompatibility leads to high maintenance costs, poor scalability, limited reusability of prior interaction patterns, and fragmented design standards, ultimately resulting in redundant development efforts~\citep{hou2025model}. To solve the issue, MCP establishes a universal and standardized framework that enables LLMs to connect with diverse tools in a seamless and secure manner, thereby facilitating coordinated utilization of various resources. 

Recently, various MCP servers with tools have been constructed in the open-source community~\citep{hou2025model}, aiming to enable LLMs to discover, select, and orchestrate different tools for real-world task-solving demands. To align with the rapid development of MCP servers, we investigate the potential of enabling \modelname{} with MCP tool-calling capabilities.

To this end, we propose \modelname{}-MCP, a model built upon \modelname{}-8B that is capable of solving a wide range of real-world tasks by interacting with various tool and data resources through MCP. We present the process of adapting \modelname{} to master MCP servers with tools. 
The overall model performance on our human-annotated MCP-tool-calling test data demonstrate the effectiveness of \modelname{}-MCP. 

\subsubsection{Data Construction}

Our data construction process consists of three main parts, including data generation, reverse data generation, and the conversion of existing function call datasets into the MCP tool using format, where reverse data generation contains single-tool and cross-tool settings. All data undergo both manual and LLM-assisted quality check procedures. More details are as follows.

\textbf{Data Generation}\quad
Many existing datasets contain high-quality queries along with annotations of the final results. We focus on identifying those queries that can be effectively addressed with the assistance of MCP servers. To this end, we manually examined the scenarios covered by publicly available datasets and selected a subset of relevant ones. For each selected dataset, we performed sampling of query completion processes using a client equipped with LLMs and MCP servers integrated in the environment. Trajectories whose final outcomes are consistent with the original annotations are retained and used for constructing training data.

\textbf{Reverse Data Generation}\quad
For the servers and tools integrated within the environment, we employ Claude-3.7-Sonnet to perform reverse query construction based on the description of each tool. Specifically, Claude-3.7-Sonnet generates queries that would necessitate the use of the target tool, and then calls the tool to solve the constructed queries. This process results in the creation of a single-tool data instance and is referred to as single-tool data reverse construction.
In addition, to train \modelname{}'s cross-tool calling capabilities, we construct cross-tool data by selecting two tools under the same server along with their respective descriptions. Claude-3.7-Sonnet then generates queries that require the simultaneous use of both tools and is constrained to call both specified tools to solve the queries.

\textbf{Conversion from Existing Tool Learning Data}\quad
We collected publicly available datasets related to tool usage and extracted data that could be parsed and converted into trajectory formats consistent with the MCP tool calling schema. These data were used to train \modelname{} with the aim of equipping it with fundamental capabilities such as adherence to tool invocation formats, accurate interpretation of user instructions, and correct filling of specified parameters. The dataset comprises approximately 140,000 instances.

All constructed data undergo a dual-phase quality inspection, involving both human evaluation and verification by a large language model (LLM). Trajectories that pass human inspection are prioritized as test data, while those that pass LLM inspection but have not been manually verified are retained as training data.

\subsubsection{Training Strategy}

We primarily adopt a learning-from-demonstration approach to train our model. The demonstrations are generated through continuous interactions between an LLM and the MCP environment. Therefore, in this section, we first introduce the key feature in building the environment. We then present the key feature of the client designed for interacting with the environment. Finally, we describe how MiniCPM learns from these demonstrations.

\textbf{MCP Environment Construction}\quad
The setup of the MCP environment involves a substantial workload and a lengthy debugging process. To minimize the effort required from developers and to enable a plug-and-play configuration, we have adopted a Docker-based approach in which all servers are installed within Docker containers. Specifically, we collect frequently used and prevalent MCP servers from both official and community MCP websites, spanning various domains such as office productivity, daily life, communication, information services, and work management. These servers are then debugged within the Docker container until they can be successfully launched.

\textbf{MCP Environment Interaction}\quad
In the MCP, the client component is responsible for interacting with the MCP servers, including performing handshakes with servers to retrieve the list of available tools. However, many MCP servers available on the market are independently deployed by community developers and have not undergone rigorous testing or quality assurance. Directly exposing all listed tools to the LLM may lead to frequent tool-calling failures, which can disrupt the execution of tasks and impair the model's performance. To address this issue, we incorporated a tool quality inspection mechanism into the client. Specifically, we prompt Claude-4 to generate 10 queries, each of which is required to invoke the current tool, based on the description and metadata of the given tool. If the execution of these queries via an LLM returns a response from the tool, regardless of whether the response is correct, this indicates that communication with the tool is functioning properly. Tools that achieve a 100\% success rate in this test are exposed to the LLM alongside their corresponding servers.


\textbf{Learning from the Demonstration}\quad
Based on the MCP Client, LLMs (e.g., \modelname{}-MCP) communicate with the Docker container, enabling seamless interaction between the LLMs and the constructed environment. We adopt a client equipped with a strong LLM to perform interaction with our constructed environment. The accumulated interaction experience is filtered to form the SFT training data of \modelname{}.

\begin{table}[t]
    \centering
    \caption{MCP tool use accuracy (\%), where ``func'', ``param'', and ``p\_v'', denote function name, parameter name, and parameter value of a tool call, respectively. ``Average'' stands for sample-weighted average accuracy across the MCP servers.}
    \resizebox{\linewidth}{!}{
    \setlength{\tabcolsep}{6pt}
    \renewcommand{\arraystretch}{1.1}
    \begin{tabular}{l ccc ccc ccc}
    \toprule
     \textbf{} & \multicolumn{3}{c}{\textbf{GPT-4o}} & \multicolumn{3}{c}{\textbf{Qwen3 8B}} & \multicolumn{3}{c}{\textbf{\modelname{}-MCP}} \\
    \cmidrule(lr){2-4} \cmidrule(lr){5-7} \cmidrule(lr){8-10}
     \textbf{MCP Servers} & \textbf{func} & \textbf{param} & \textbf{p\_v} & \textbf{func} & \textbf{param} & \textbf{p\_v} & \textbf{func} & \textbf{param} & \textbf{p\_v} \\
    \midrule
    Airbnb & 89.3 & 67.9 & 53.6 & 92.8 & 60.7 & 50.0 & 96.4 & 67.9 & 50.0 \\
    Amap-Maps & 79.8 & 77.5 & 50.0 & 74.4 & 72.0 & 41.0 & 89.3 & 85.7 & 39.9 \\
    Arxiv-MCP-Server & 85.7 & 85.7 & 85.7 & 81.8 & 54.5 & 50.0 & 57.1 & 57.1 & 52.4 \\
    Calculator & 100.0 & 100.0 & 20.0 & 80.0 & 80.0 & 13.3 & 100.0 & 100.0 & 6.67 \\
    Computor-Control-MCP & 90.0 & 90.0 & 90.0 & 90.0 & 90.0 & 90.0 & 90.0 & 90.0 & 86.7 \\
    Desktop-Commander & 100.0 & 100.0 & 100.0 & 100.0 & 100.0 & 100.0 & 100.0 & 100.0 & 100.0 \\
    Filesystem & 63.5 & 63.5 & 31.3 & 69.7 & 69.7 & 26.0 & 83.3 & 83.3 & 42.7 \\
    Github & 92.0 & 80.0 & 58.0 & 80.5 & 50.0 & 27.7 & 62.8 & 25.7 & 17.1 \\
    Gaode & 71.1 & 55.6 & 17.8 & 68.8 & 46.6 & 24.4 & 68.9 & 46.7 & 15.6 \\
    MCP-Code-Executor & 85.0 & 80.0 & 70.0 & 80.0 & 80.0 & 70.0 & 90.0 & 90.0 & 65.0 \\
    MCP-Docx & 95.8 & 86.7 & 67.1 & 94.9 & 81.6 & 60.1 & 95.1 & 86.6 & 76.1 \\
    PPT & 72.6 & 49.8 & 40.9 & 85.9 & 50.7 & 37.5 & 91.2 & 72.1 & 56.7 \\
    PPTx & 64.2 & 53.7 & 13.4 & 91.0 & 68.6 & 20.9 & 91.0 & 58.2 & 26.9 \\
    Simple-Time-Server & 90.0 & 70.0 & 70.0 & 90.0 & 90.0 & 90.0 & 90.0 & 60.0 & 60.0 \\
    Slack & 100.0 & 90.0 & 70.0 & 100.0 & 100.0 & 65.0 & 100.0 & 100.0 & 100.0 \\
    Whisper & 90.0 & 90.0 & 90.0 & 90.0 & 90.0 & 90.0 & 90.0 & 90.0 & 30.0 \\
    \midrule
    Average & 80.2 & 70.2 & 49.1 & 83.5 & 67.7 & 43.8 & \textbf{88.3} & \textbf{76.1} & \textbf{51.2}  \\
    \bottomrule
    \end{tabular}}
    \label{mcp:overall-results}
\end{table}

\subsubsection{Evaluation}

\textbf{Evaluation Details}\quad
Following the existing commonly adopted evaluation metrics~\citep{qin2024toolllm}, we evaluate the accuracy of the tool name, parameter names, and parameter values for each tool call in our human-annotated test data. For those traces containing multiple steps of tool calls, we evaluate the accuracy of the current step by giving the previous ground-truth steps.

\textbf{Results}\quad
The overall results of model performance on our human-annotated MCP-tool-calling test data are shown in Table~\ref{mcp:overall-results}. According to the experimental results, we find that Qwen3-8B possesses the basic MCP Tool calling capability, demonstrating a competent understanding and usage of MCP tools, primarily because the experience of invoking MCP tools overlaps significantly with that of invoking regular tools. However, when it comes to newer tools or tools in more specialized domains (e.g., arXiv, airbnb, etc.), Qwen3 appears less familiar. It tends to apply prior knowledge from other tools when generating parameter names and passing parameter values, without adequately adjusting or adapting to the specific requirements of the given MCP tool. In comparison, \modelname{} learns from the demonstrations and thus knows the characteristics of our collected MCP servers and tools, which leads to a better performance on the test data.

\section{Conclusion and Future Works}
In this technical report, we present \modelname{}, which features efficient pre-training and inference. Thanks to the efficient pre-training data and infrastructure, we can use only 8 trillion tokens to reach comparable performance with existing open-source models. And the efficient architecture and inference systems, we can achieve $5\times$ speedup for long-sequence processing. To facilitate the development of open-source community, we release the model parameters and inference code of \modelname{}. 

In the future, we will continually investigate efficient training and inference of LLMs. 
In terms of model architecture, we will devote our effort to the efficient sparse model architecture with the target to enable LLMs process infinitely long sequences on end-side devices. In terms of data construction, our research will focus on improving the quality of existing corpus and synthesise large-scale reasoning-intensive pre-training datasets, which can significantly improve the foundational capabilities of LLMs. In addition, we will continually explore the great potential of reinforcement learning to enable LLMs to learn skills from various environments. As for the inference systems, we plan to develop efficient systems for most end-side platforms, which can help the community to run and evaluate our model.

\section{Contributions and Acknowledgments}
\modelname{} and \modelname{}.1 are the result of the collective efforts of all members of our team. 

\textbf{Project Design and Coordination}\quad
Chaojun Xiao, Yuxuan Li,  Xu Han

\textbf{Contributors} (Ordered by the last name)\quad
Yuzhuo Bai, Jie Cai, Haotian Chen, Wentong Chen, Xin Cong, Ganqu Cui,
Ning Ding,
Shengda Fan, Yewei Fang, Zixuan Fu,
Wenyu Guan, Yitong Guan, Junshao Guo,
Yufeng Han, Bingxiang He, Yuxiang Huang,
Baoxi Ji,
Cunliang Kong,
Qiuzuo Li, Siyuan Li, Wenhao Li, Xin Li, Yanghao Li, Yishan Li, Zhen Li, Dan Liu, Biyuan Lin, Yankai Lin, Xiang Long, Quanyu Lu, Yaxi Lu, Peiyan Luo, Hongya Lyu,
Litu Ou,
Yinxu Pan, Lushi Pu,
Zekai Qu,
Qundong Shi, Zijun Song, Jiayuan Su, Zhou Su, Ao Sun, Xianghui Sun,
Peijun Tang,
Fangzheng Wang, Feng Wang, Shuo Wang, Yudong Wang, Zheng Wang, Yesai Wu,
Zhenyu Xiao, Jie Xie, Zihao Xie, Xiaoyue Xu,
Yukun Yan, Jiarui Yuan,
Jinqian Zhang, Kaihuo Zhang, Lei Zhang, Linyue Zhang, Xueren Zhang, Yudi Zhang, Hengyu Zhao, Weilin Zhao, Weilun Zhao, Yuanqian Zhao, Zhi Zheng, Chuyue Zhou, Ge Zhou, Jie Zhou, Wei Zhou, Yanghao Zhou, Zihan Zhou, Zixuan Zhou

\textbf{Supervision}\quad
Xu Han, Zhiyuan Liu, Guoyang Zeng, Chao Jia, Dahai Li, Maosong Sun

\newpage

\bibliographystyle{citation}
\bibliography{citation}


\end{document}